\documentclass[10pt,journal,twoside,twocolumn,final]{IEEEtran}
\usepackage{cite}
\usepackage[pdftex]{graphicx}
\usepackage{amsmath}
\usepackage{algorithm,algorithmic}
\usepackage{array}
\usepackage[caption=false,font=footnotesize,labelfont=sf,textfont=sf]{subfig}
\usepackage{threeparttable}
\usepackage[utf8]{inputenc} 
\usepackage[T1]{fontenc}    
\usepackage{hyperref}       
\usepackage{url}            
\usepackage{booktabs}       
\usepackage{amsfonts}       
\usepackage{nicefrac}       
\usepackage{microtype}      
\usepackage{enumerate}
\usepackage{multirow}
\usepackage{xcolor}
\usepackage{bm}
\usepackage{bbm}
\usepackage{color}
\newtheorem{mypro}{Proposition}
\newtheorem{lemma}{Lemma}

\begin{document}

\title{NCGNN: Node-Level Capsule Graph Neural Network for Semisupervised Classification}

\author{
 Rui~Yang,
 Wenrui~Dai,~\IEEEmembership{Member,~IEEE,}
 Chenglin~Li,~\IEEEmembership{Member,~IEEE,}
 Junni~Zou,~\IEEEmembership{Member,~IEEE,}
 and~Hongkai~Xiong,~\IEEEmembership{Senior Member,~IEEE}
 \thanks{This work was supported in part by the National Natural Science Foundation of China under Grant 61932022, Grant 61931023, Grant 61971285, Grant 61831018, Grant 61871267, Grant 61720106001, Grant 62120106007, Grant 61972256, Grant T2122024, Grant 62125109; in part by the Program of Shanghai Science and Technology Innovation Project under Grant 20511100100; and in part by Central Hardware Engineering Institute of Huawei. (\emph{Corresponding author: Wenrui Dai.})}
 \thanks{Rui~Yang, Chenglin~Li and Hongkai~Xiong are with the Department of Electronic Engineering, Shanghai Jiao Tong University, Shanghai 200240, China (e-mail: rui\_yang@sjtu.edu.cn, lcl1985@sjtu.edu.cn, xionghongkai@sjtu.edu.cn).}
 \thanks{Wenrui~Dai and Junni~Zou are with the Department of Computer Science and Engineering, Shanghai Jiao Tong University, Shanghai 200240, China (e-mail: daiwenrui@sjtu.edu.cn, zoujunni@sjtu.edu.cn).}
}

\markboth{IEEE Transactions on Neural Networks and Learning Systems}
{Yang \MakeLowercase{\textit{et al.}}: NCGNN: Node-Level Capsule Graph Neural Network for Semisupervised Classification}

\maketitle

\begin{abstract}
Message passing has evolved as an effective tool for designing Graph Neural Networks (GNNs). However, most existing methods for message passing simply sum or average all the neighboring features to update node representations. They are restricted by two problems, i.e., (i) lack of interpretability to identify node features significant to the prediction of GNNs, and (ii) feature over-mixing that leads to the over-smoothing issue in capturing long-range dependencies and inability to handle graphs under heterophily or low homophily. In this paper, we propose a Node-level Capsule Graph Neural Network (NCGNN) to address these problems with an improved message passing scheme. Specifically, NCGNN represents nodes as groups of node-level capsules, in which each capsule extracts distinctive features of its corresponding node. For each node-level capsule, a novel dynamic routing procedure is developed to adaptively select appropriate capsules for aggregation from a subgraph identified by the designed graph filter. NCGNN aggregates only the advantageous capsules and restrains irrelevant messages to avoid over-mixing features of interacting nodes. Therefore, it can relieve the over-smoothing issue and learn effective node representations over graphs with homophily or heterophily. Furthermore, our proposed message passing scheme is inherently interpretable and exempt from complex post-hoc explanations, as the graph filter and the dynamic routing procedure identify a subset of node features that are most significant to the model prediction from the extracted subgraph. Extensive experiments on synthetic as well as real-world graphs demonstrate that NCGNN can well address the over-smoothing issue and produce better node representations for semisupervised node classification. It outperforms the state of the arts under both homophily and heterophily.
\end{abstract}
	
\begin{IEEEkeywords}
Semisupervised classification, Graph Neural Networks (GNNs), message passing, capsule, over-smoothing, homophily and heterophily.
\end{IEEEkeywords}

\section{Introduction}
\IEEEPARstart{M}{essage} passing has proved to be the cornerstone of numerous Graph Neural Networks (GNNs) for analyzing data with irregular structures, including citation networks~\cite{gcn,lgcn,gat}, point clouds~\cite{ecc,edgeconv}, chemical molecules~\cite{molecule,mpnn}, and recommendation systems~\cite{recom}. The essential idea is to update the representation of each node by aggregating features from its topological neighborhoods. Commencing with the great success of GCN~\cite{gcn} in semisupervised node classification, a myriad of studies have been developed to ameliorate the vanilla message passing scheme with attention mechanism~\cite{gat,finger}, edge attributes~\cite{edge}, multi-relational features~\cite{relation}, structural information~\cite{geomgcn,curvature,finger}, etc.

Despite the promising performance in many graph-based machine learning tasks, most existing message passing schemes adopt the naive sum or average aggregator to fuse all the neighborhood information, and are designed under the assumption of strong homophily, i.e., ``birds of a feather flock together''~\cite{birds}. Therefore, challenges have risen for GNNs to solve semisupervised node classification, as summarized below.
\begin{itemize}
\item \textbf{Interpretability:} Summing up or averaging all the neighboring features makes GNNs lack human intelligible explanations. Since the most influential part of node features for the prediction of the model cannot be identified, complex post-hoc methods would be required to explain GNNs in certain application domains (e.g., GNNExplainer~\cite{explainer} and XGNN~\cite{xgnn}).
\item \textbf{Over-smoothing:} Excessive irrelevant messages (noise) might be aggregated, when stacking multiple message passing layers to capture long-range (high-order) neighborhood information. Node features in different classes would be over-mixed and indistinguishable, which is the over-smoothing issue in GNNs.
\item \textbf{Heterophily:} The naive average aggregator is essentially a special form of Laplacian smoothing~\cite{deeper} that forces proximal nodes to have similar features. This working mechanism is desirable for graphs under the assumption of strong homophily (e.g., citation networks), but fails to generalize to networks with heterophily or low homophily, where connected nodes are more likely from different communities (classes) and have dissimilar features (e.g., amino acids in protein structures~\cite{protein}). Feature averaging over the heterophily-dominant neighborhoods would also cause feature over-mixing that produces undesirable similar features for nodes in different classes. 
\end{itemize}

Inspired by the interpretability and dynamic routing procedure of capsule networks~\cite{caps4,caps1,caps2,caps_equi,caps3}, several capsule-related GNNs~\cite{capsule2,capsule0,capsule1,disen,caps2ne,hgcn} have been developed in the graph domain. However, they commonly focus on \emph{graph-level} classification or flatten the capsules back to conventional scalar features during message passing. In this paper, we propose a Node-level Capsule Graph Neural Network (NCGNN) to address the aforementioned challenges with adaptive message passing between interacting nodes. The contributions of this paper are summarized as below. 
\begin{itemize}
\item We propose vectorized node-level capsule (a group of neurons) for feature aggregation to efficiently preserve the distinctive properties of nodes.
\item We develop a novel dynamic routing procedure that adaptively selects appropriate node-level capsules for aggregation to prevent feature over-mixing and generate interpretable high-level capsules for each class.
\item We demonstrate in theory that the proposed message passing scheme is permutation invariant and improves the expressive power of vanilla message passing scheme.
\end{itemize}

To the best of our knowledge, NCGNN is the first GNN model that leverages inherently interpretable \emph{node-level} capsule-based message passing to improve semisupervised node classification performance. In the existing work, DisenGCN~\cite{disen} is the only capsule-related GNN for node classification, where message passing is realized based on the multi-head mechanism shown in Fig.~\ref{fig1}(b). By comparison, NCGNN exploits dynamic routing between node-level capsules to adaptively optimize the selection of subset of features from partial graph nodes (subgraph) and produce better node representations, as depicted in Fig.~\ref{fig1}(c). With such ability to overcome feature over-mixing in neighborhood feature aggregation, NCGNN goes beyond the homophily limitation and relieves the over-smoothing problem for learning from larger receptive field.

To be concrete, NCGNN first transforms each node from the scalar-based features to a group of primary node capsules (vectors). Node features are projected into different subspaces such that each capsule reflects the distinctive properties of a node. Subsequently, for each target node, we employ two graph filters to determine a subgraph of neighboring nodes for feature aggregation and propose a novel neighborhood-routing-by-agreement mechanism that adaptively selects advantageous node-level capsules to generate class node capsules for classification. Therefore, the node-level capsule-based message passing is inherently interpretable by identifying a subset of input node capsules that are most significant to the model prediction from the extracted subgraph.

For empirical evaluations, we conduct comprehensive experiments on twelve real-world graphs with homophily or herterophily and synthetic graphs where the homophily level can be controlled. Experimental results demonstrate that, contrary to existing message passing schemes that might suffer from the over-smoothing issue in extracting high-order neighborhood information and perform poorly under the heterophily setting, NCGNN achieves performance gains with the increasing orders (hops) of neighborhoods and yields state-of-the-art performance under both homophily and heterophily. We further show that the dynamic routing procedure between node-level capsules indeed produces more separable node embeddings for nodes in different classes, which also verifies the effectiveness of NCGNN for tackling the over-smoothing problem.

The rest of this paper is organized as follows. Section~\ref{sec2} reviews the background and related work. Section~\ref{sec3} elaborates the proposed NCGNN framework. Experimental results are provided in Section~\ref{sec4} and conclusions are drawn in Section~\ref{sec5}.

\section{Background and Related Work} \label{sec2}
\subsection{Message Passing Based GNNs} \label{message}
The core of most GNNs is message passing, in which each node aggregates features from its topological neighborhoods to update its representation. Let us denote the learnable weights matrix as $\mathbf{W}\in\mathbb{R}^{{f_o}\times {f_i}}$, the set of neighboring node indices for node $v_i$ as $\mathcal{N}(i)$ , and the nonlinear activation function (e.g., ReLU) as $\sigma(\cdot)$. A basic message passing layer takes the following form:
\begin{align} \label{eq1}
    \mathbf{h}_{i}^{\prime} = \sigma\left(\sum\nolimits_{j\in \mathcal{N}(i)\cup\{i\}}\alpha_{ij}\mathbf{W}\mathbf{h}_{j}\right),
\end{align}
where $\mathbf{h}_{j}\in\mathbb{R}^{f_i}$ is the input features of node $v_j$, $\mathbf{h}_{i}^{\prime}\in\mathbb{R}^{f_o}$ is the output features of node $v_i$, $\mathbf{W}\mathbf{h}_{j}\in\mathbb{R}^{f_o}$ is the transformed neural message sent from node $v_j$, and $\alpha_{ij}$ is the edge weight for the message passing from node $v_j$ to node $v_i$, which can be computed via different mechanisms~\cite{gcn,gat,disen,curvature,finger}.

\begin{figure}[t]
\centerline{\includegraphics[scale=0.5]{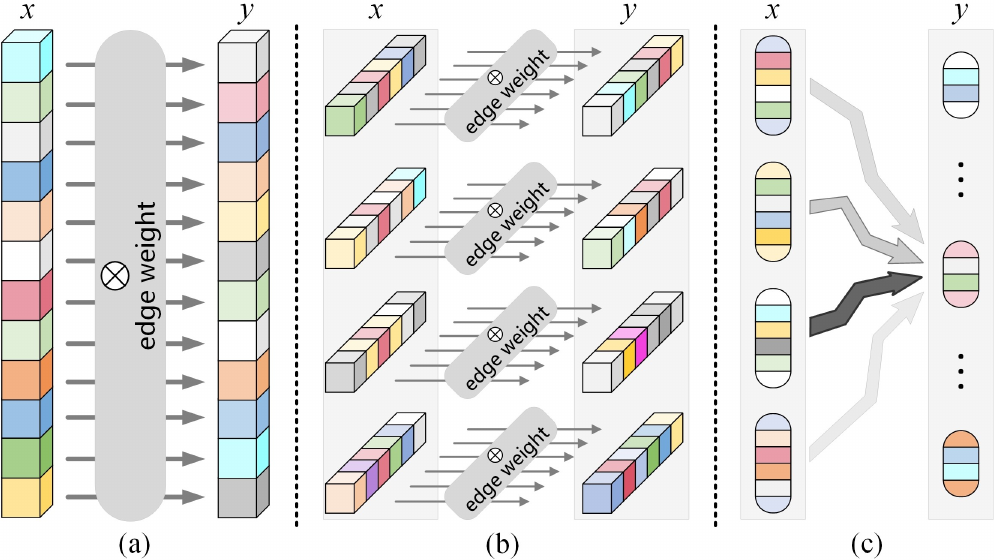}}
\caption{Comparison of different message passing schemes from node $x$ to node $y$. Each cubic is a scalar while each capsule denotes a vector. (a) The vanilla sum aggregator. (b) The multi-channel message passing scheme with four sum aggregators. (c) The dynamic routing procedure between node-level capsules in NCGNN. For a better view, we only plot the dynamic routing between one capsule of node $y$ and the capsules of node $x$. The intensity of the arrow color indicates the coupling degree between node-level capsules.}
\label{fig1}
\end{figure}

Fig.~\ref{fig1} provides a brief comparison of different message passing schemes. Fig.~\ref{fig1}(a) depicts the vanilla sum aggregator in GCN~\cite{gcn}, where each node sums up all the neighboring features with fixed edge weights to update its representation. Fig.~\ref{fig1}(b) shows the multi-channel attention utilized by models such as GAT~\cite{gat} and DisenGCN~\cite{disen}. Although the edge weights can be learned with diverse mechanisms to improve the model capacity, each channel is still a naive averaging operation that sums up all the neighboring features, which can be viewed as an ensemble of GNN models. Fig.~\ref{fig1}(c) demonstrates the dynamic routing procedure in our NCGNN, where each node-level capsule adaptively aggregates an appropriate subset of capsules and restrains irrelevant messages from neighboring nodes to prevent feature over-mixing, thereby generating better node representations for the downstream classification.

\subsection{Homophily and Heterophily}
Graph-structured data are ubiquitous in a diverse set of domains. In this paper, we categorize graphs into assortative and disassortative ones according to the homophily/heterophily level of node class labels. As in~\cite{geomgcn,beyond}, the metric to measure the homophily ratio of graph $\mathcal{G}$ is defined as:
\begin{align} \label{eq2}
    H(\mathcal{G}) = \frac{1}{|\mathcal{V}|}\sum_{v_i\in\mathcal{V}}\frac{\sum_{j\in\mathcal{N}(i)}\mathbbm{1}(y_i=y_j)}{|\mathcal{N}(i)|},
\end{align}
where $\mathcal{V}$ is the node set, $y_i$ and $y_j$ denote the labels of nodes $v_i$ and $v_j$, $\mathbbm{1}(x) = 1$ if $x$ is true and 0 otherwise.

Assortative graphs are those with high homophily ratios (i.e., $H(\mathcal{G})$ is closer to one) such as citation networks and social networks. On the other hand, disassortative graphs exhibit heterophily or low homophily property (i.e., $H(\mathcal{G})$ is closer to zero). Connections of different types of amino acids in protein structures and web-page linking networks are two typical examples of disassortative graphs.

Many existing message passing schemes described by Eq.~\eqref{eq1} rely on the strong homophily assumption and most common benchmark datasets in the task of node classification adhere to this principle (e.g., Cora~\cite{collective}, Citeseer~\cite{collective}, and Pubmed~\cite{query}). In this scenerio, feature averaging over each node's neighborhood would produce similar representations for nodes in the same class, which makes the subsequent classification task much easier. However, they fail to generalize to the heterophily setting and are even worse than the models that ignore the graph structure (e.g., Multi-Layer Perceptron or MLP)~\cite{beyond}, since summing up all the neighboring features from different communities would aggregate too much noise, thus over-mixing and generating inseparable features for nodes in different classes. Several works have been proposed to address this limitation. Geom-GCN~\cite{geomgcn} builds structural neighborhoods in a latent space and proposes a novel bi-level aggregation scheme based on the observation that nodes of the same class in some disassortative graphs exhibit high structural similarity~\cite{struc2vec}. $\text{H}_{\text{2}}\text{GCN}$~\cite{beyond} identifies a set of key designs that improves learning from disassortative graphs: ego- and neighbor-embedding separation, higher-order neighborhoods, and combination of intermediate representations. Different from these methods, we propose to use dynamic routing between node-level capsules to adaptively aggregate a subset of appropriate capsules from a selected subgraph. Thus, irrelevent messages can be effectively restrained for learning under both homophily and heterophily.

\subsection{Over-smoothing in GNNs}
As illustrated in~\cite{measuring}, the over-smoothing issue is caused by the feature over-mixing of nodes in different classes when enlarging the receptive field to capture high-order (multi-hop) neighborhood information. To address this problem, numerous works have been proposed in recent years. JK-Net~\cite{jknet} and DNA~\cite{dna} utilize skip connection and attention mechanism for preserving the information in each message passing step. PPNP and APPNP~\cite{pagerank} consider graph diffusion to replace the adjacency matrix with personalized PageRank matrix that takes the root node or initial node features into account. In~\cite{deepgcns}, GCNs are armed with residual/dense connections. GCNII~\cite{gcnii} modifies GCN with initial residual and identity mapping. DAGNN~\cite{dagnn} decouples feature transformation and propagation and proposes a novel adjustment mechanism to adaptively aggregate features of different propagation hops. Scattering GCN~\cite{scattering} uses geometric scattering transform to enable band-pass filtering over the graph signal. PairNorm~\cite{pairnorm} and Differentiable Group Normalization (DGN)~\cite{dgn} are two normalization schemes proposed for GNNs with deeper layers. Besides, some tricks have been studied. DropEdge~\cite{dropedge} randomly removes a certain fraction of edges in each training iteration to reduce message passing. AdaEdge~\cite{measuring} constantly adjusts the graph topology (adding intra-class edges and removing inter-class edges) according to the predictions made by GNNs in a self-training-like fashion. BBGDC~\cite{bbgdc} further generalizes DropEdge and Dropout~\cite{dropout} by adaptive connection sampling. In comparison, NCGNN relieves the over-smoothing issue by directly restraining noisy messages passing between interacting nodes to prevent feature over-mixing.

\subsection{Interpretability Methods for GNNs}
Designing explainable approaches for GNNs is an emerging research field. Most of existing interpretability methods adopt the post-hoc explanation for a trained GNN. In~\cite{explainability1}, several explanation techniques that are originally used for standard neural networks are applied to GNNs, such as Sensitivity Analysis~\cite{sa}, Guided Backpropagation~\cite{gbp}, and Layer-wise Relevance Propagation~\cite{lrp}. Gradient-based saliency maps~\cite{saliency}, Class Activation Mapping~\cite{cam}, and Excitation Backpropagation~\cite{excitationbackprop} are extended from conventional Convolutional Neural Networks (CNNs) to GNNs in~\cite{explainability2}. GNNExplainer~\cite{explainer} generates explanation by identifying a subgraph and a subset of node features that are crucial for the model’s prediction of each instance. PGM-Explainer~\cite{pgm} further resorts to probabilistic graphical model to capture the dependencies among explained features. PGExplainer~\cite{pgexplainer} uses parameterized neural network to identify the important subgraphs at the model-level. XGNN~\cite{xgnn} and RG-Explainer~\cite{rgexplainer} generate explanatory graph with reinforcement learning at the model-level and instance-level, respectively. GNN-LRP~\cite{xai} identifies relevant walks in the input graph to interpret GNN's prediction. More recently, SubgraphX~\cite{subgraphx} explores significant subgraphs for explaining GNNs with Monte Carlo tree search. Gem~\cite{gem}, CF-GNNExplainer~\cite{cf}, and RCExplainer~\cite{rcfgnn} provide explanations through causal inference. ReFine~\cite{mgegnn} approaches multi-grained interpretability that provides both model- and instance-level explanations. And ie-HGCN~\cite{iehgcn} focuses on interpreting heterogeneous-information-network-oriented GCNs via automatically learning useful meta-paths. Contrary to most of the above interpretability methods that provide a post-hoc explanatory subgraph at the \emph{structure-level} (e.g., PGExplainer~\cite{pgexplainer}, SubgraphX~\cite{subgraphx}, and Gem~\cite{gem}), we design an inherently interpretable message passing scheme for homogeneous graphs at the \emph{feature-level} by identifying important input node capsules from the extracted subgraph.

\begin{figure*}[t]
\centerline{\includegraphics[scale=0.65]{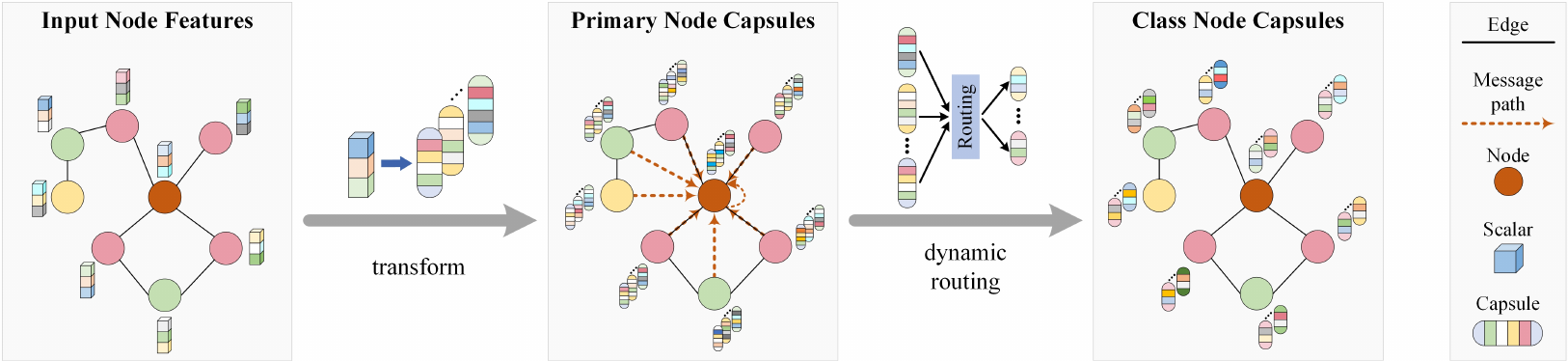}}
\caption{Workflow of NCGNN: NCGNN first transforms each node from the scalar-based features to $K$ primary node capsules. Subsequently, the designed graph filter selects a subgraph for each target node to perform message passing and dynamic routing is leveraged to generate $C$ class node capsules for it, where $C$ is the number of classes. In capsule graph layer, each node-level capsule adaptively selects appropriate capsules to update its representation via dynamic routing and irrelevant messages from neighborhoods are restrained to avoid feature over-mixing. For a better view, we only plot the message passing procedure for generating the class node capsules of the brown node within its ego network of subgraph. Note that capsule graph layers can be further stacked if necessary, though we only use one capsule graph layer to produce $C$ class node capsules for each target node.}
\label{fig2}
\end{figure*}

\subsection{Capsule-related Networks}\label{capsule}
The concept of capsule was first put forward in \cite{caps4}, and recently it has achieved enormous advances (e.g.,~\cite{caps1,caps2,caps_equi,caps3,inter_caps}). The key is to replace the scalar-based features in conventional CNNs with vector-based capsules to extract high-quality image representations. In CapsNet~\cite{caps1}, each capsule is a group of neurons whose overall length represents the existence of an entity and its orientation encapsulates the instantiation parameters. The capsules in the last layer are called class capsules, and the length of each  class capsule's output vector indicates the predicted probability of each class. The advantage is that it utilizes dynamic routing to determine the coupling degree between every two capsules in successive layers, thereby benefiting the information propagation in the network to select appropriate capsules for generating higher-level representations. Capsule-related networks have good nature of interpretability since the instantiation parameters encode various types of properties and the capsule selection mechanism of dynamic routing can identify crucial features for the network output, which helps explain the model prediction.

Capsules have been extended to the graph domain. GCAPS-CNN~\cite{capsule2} concatenates higher-order statistical moments of each node's neighboring features as its capsules. However, capsules are flattened back to scalar features during message passing. In~\cite{capsule0}, explicit kernel graph representations and CapsNet are combined to perform classification. CapsGNN~\cite{capsule1} employs a GNN to produce graph-level representations and utilizes dynamic routing on top of the GNN to benefit graph embeddings. HGCN~\cite{hgcn} further utilizes capsule to characterize part-whole relationship for learning hierarchical graph representations. All of the above methods focus on learning more effective graph-level embeddings for graph classification, while our work is proposed for semisupervised node classification. As for node-level learning, DisenGCN~\cite{disen} views the disentangled node representations as capsules. However, its message passing scheme is essentially a multi-channel ensemble as shown in Fig.~\ref{fig1}(b). There is no dynamic routing between different channels (capsules) and capsules are also flattened back to scalar-based features like GCAPS-CNN. By contrast, NCGNN leverages dynamic routing to adaptively aggregate appropriate node-level capsules and restrain harmful noise from higher-order neighborhoods for each target node. Caps2NE~\cite{caps2ne} treats each node as one capsule for unsupervised node representation learning, while NCGNN treats each node as a group of capsules for semisupervised node classification.

\section{Node-level Capsule Graph Neural Network} \label{sec3}
In this section, we elaborate the proposed Node-level Capsule Graph Neural Network (NCGNN) for semisupervised node classification. NCGNN has good inherent interpretability with an adaptive message passing scheme that represents each node as a group of capsules and leverages dynamic routing between capsules. NCGNN is capable of adaptively identifying a subset of crucial node-level capsules during the message passing procedure. Therefore, NCGNN can help relieve the over-smoothing issue and learn effectively under both homophily and heterophily by avoiding feature over-mixing. Fig.~\ref{fig2} demonstrates the workflow of NCGNN.

\subsection{Problem Formulation}
We focus on the task of graph-based semisupervised node classification. An undirected graph with $N$ nodes is represented as a tuple $\mathcal{G}=(\mathcal{V},\mathbf{A})$, where $\mathcal{V}=\{v_i\}_{i=1}^{N}$ is the node set, and $\mathbf{A} \in \mathbb{R}^{N \times N}$ is the adjacency matrix where each entry $a_{ij}$ is set to 1 for the existence of edge linking nodes $v_{i}$ and $v_{j}$ and 0 otherwise. The normalized graph Laplacian is defined as $\mathbf{L}=\mathbf{I}_{N}-\mathbf{D}^{-\frac{1}{2}}\mathbf{A}\mathbf{D}^{-\frac{1}{2}}$, where $\mathbf{I}_{N}$ is the identity matrix and $\mathbf{D}=\operatorname{diag}(d_1,\ldots,d_N)$ is the degree matrix with $d_i$ denoting the degree of node $v_i$. The eigendecomposition of $\mathbf{L}$ gives us $\mathbf{L}=\mathbf{U}\mathbf{\Lambda}\mathbf{U}^\top$. Here, $\mathbf{\Lambda}$ is a diagonal matrix of the eigenvalues of $\mathbf{L}$, and the columns of $\mathbf{U}$ are the orthonormal eigenvectors termed graph Fourier basis. According to the spectral graph theory, a filter on $\mathcal{G}$ can be defined as a parameterized polynomial function $g_{\theta}(\mathbf{L})$ of $\mathbf{L}$, and a graph signal $x\in \mathbb{R}^{N}$ filtered by $g_{\theta}$ can be written as $y=g_{\theta}(\mathbf{L})x = \mathbf{U}g_{\theta}(\mathbf{\Lambda})\mathbf{U}^\top x \in \mathbb{R}^{N}$. We also have a feature matrix $\mathbf{X}=[\mathbf{x}_1,\ldots,\mathbf{x}_N]^\top \in \mathbb{R}^{N \times f}$ with each row corresponding to the feature vector of an individual node. Given a subset of graph nodes $\mathcal{V}_l \subset{\mathcal{V}}$ with labels, the task of semisupervised node classification is to predict the classes of unlabeled nodes $\mathcal{V}_u=\mathcal{V}\backslash\mathcal{V}_l$ based on $\mathcal{G}$ and $\mathbf{X}$.

\subsection{Motivations}
In most existing GNNs~\cite{gcn,monet,gat}, message passing is realized by summing up or averaging the node features within a neighborhood. They cannot distinguish the nodes and their features that would lead to feature over-mixing in certain scenarios such as heterophily and larger receptive field, and consequently, harm the downstream node classification. Therefore, we design a message passing scheme for NCGNN to address two enlightening questions as below. 
\begin{itemize}
\item \textbf{Q1}: For each target node, how to identify the graph nodes (a subgraph) whose information should be aggregated?
\item \textbf{Q2}: For each target node-level capsule, how to determine and aggregate an appropriate subset of capsules that contains little noise from the subgraph?
\end{itemize}

First, we address \textbf{Q1} by selecting a subgraph for each target node with two graph filters designed in the attention- and diffusion-based manner, respectively. Contrary to most existing GNNs (e.g., GCN \cite{gcn}, MoNet \cite{monet}, and GAT \cite{gat}) that only aggregate one-hop neighborhood information in each message passing step and stack multiple layers with entangled feature transformation and propagation to enlarge the receptive field, the designed graph filters can adapt to higher-order neighborhoods to directly capture long-range dependencies without increasing the model complexity. Therefore, NCGNN can reduce the risk of over-smoothing by decoupling feature aggregation and transformation~\cite{dagnn} and enables explanations from both spatial and spectral aspects.

Moreover, we solve \textbf{Q2} by designing a capsule graph layer where message passing is improved with a novel neighborhood-routing-by-agreement mechanism between node-level capsules. As mentioned above, features from neighboring nodes in real-world graphs are not always desirable, especially when the receptive field is enlarged to extract long-range dependencies and tackling networks with heterophily or low homophily. Under these circumstances, averaging over all the neighboring features would aggregate excessive irrelevant messages and over-mix node features from different classes, which severely hampers the node classification performance. By contrast, the proposed dynamic routing procedure adaptively selects and aggregates the advantageous neighboring capsules for each node-level capsule in our capsule graph layer. Harmful noise from neighborhoods would be restrained to relieve the over-smoothing issue and node representations could be learned effectively under both homophily and heterophily.

\subsection{Primary Node Capsules}
NCGNN first converts each node from the scalar-based features to a group of primary node capsules by projecting the node features into $K$ different subspaces, followed by the normalization operation, which can take the following form:
\begin{align} \label{eq3}
    \mathbf{h}_{i}^{(k)} = \frac{\sigma(\dot{\mathbf{W}}_k\mathbf{x}_{i} + \dot{\mathbf{b}}_k)}{\|\sigma(\dot{\mathbf{W}}_k\mathbf{x}_{i} + \dot{\mathbf{b}}_k)\|}, \ \ k=1,\ldots,K,
\end{align}
where $\|\cdot\|$ denotes the Euclidean norm, $\mathbf{h}_{i}^{(k)}\in\mathbb{R}^{f_p}$ is the $k$-th primary node capsule of node $v_i$ that describes the aspect related to the $k$-th subspace, $\dot{\mathbf{W}}_k\in\mathbb{R}^{f_p\times f}$ and $\dot{\mathbf{b}}_k\in\mathbb{R}^{f_p}$ are the learnable weights and bias with respect to the $k$-th subspace, and $f_p$ is the dimension of each subspace. Note that although some more sophisticated implementations can be applied to further enhance the informativeness for each subspace such as enforcing the primary node capsules of each node to be independent with Hilbert-Schmidt Independence Criterion (HSIC)~\cite{independence}, we find that the simple linear projection still achieves remarkable performance in practice. Thus, we use Eq.~\eqref{eq3} to generate $K$ primary node capsules for each graph node considering the efficiency issue.

Having obtained the primary node capsules, in what follows we elaborate the mechanisms of the designed graph filter and capsule graph layer to solve questions \textbf{Q1} and \textbf{Q2} for producing better node representations.

\subsection{Graph Filter}
For each target node, the graph filter identifies a subgraph where information should be aggregated and determines the influence of each subgraph node on it. In the vanilla GCN~\cite{gcn}, the graph filter is defined as the symmetric normalized transition matrix $\Tilde{\mathbf{A}}=\hat{\mathbf{D}}^{-\frac{1}{2}}\hat{\mathbf{A}}\hat{\mathbf{D}}^{-\frac{1}{2}}$, where $\hat{\mathbf{A}}=\mathbf{A}+\mathbf{I}_N$ and $\hat{\mathbf{D}}=\mathbf{D}+\mathbf{I}_N$. This graph filter and its variants~\cite{gat,finger} are widely used in existing GNNs where only the one-hop neighborhood features are aggregated. However, higher-order neighborhood information is also useful for generating node representations, as studied in~\cite{mixhop,motif,pagerank}. To extract long-range dependencies, we define the generalized graph filter as in Eq.~\eqref{eq4} with two specific forms in the attention- and diffusion-based manner, respectively.
\begin{align} \label{eq4}
\bar{\mathbf{A}}=f_{\xi}(\tilde{\mathbf{A}})=\sum_{i\in \mathcal{M}, \ \mathcal{M} \subseteq{\mathbb{N}}}\xi_{i}\tilde{\mathbf{A}}^{i}.
\end{align}
Here, $\mathcal{M}$ is the set of multiple orders (hops) (e.g., $\mathcal{M}=\{1,2,4,5\}$), $f_{\xi}$ is a polynomial function parameterized by coefficients $\xi_{i}$ ($i\in\mathcal{M}$). Considering that the exponentiation operation $\tilde{\mathbf{A}}^{i}$ may result in a dense matrix even for a small $i$ due to the small-world property~\cite{smallworld} of real-world graphs, we maintain the sparsity property by using either an $\epsilon$-threshold or top-$k$ sparsification as in~\cite{gdc}, i.e., entries of $\tilde{\mathbf{A}}^{i}$ that are smaller than a threshold or are not the top-$k$ values in each row are set to 0, and the symmetric normalization are recalculated for the sparsified version of $\tilde{\mathbf{A}}^{i}$ subsequently. Then, each row of $\bar{\mathbf{A}}$ identifies part of the $N$ nodes (a subgraph) for the corresponding node to aggregate information, and each non-zero entry denotes the influence of the message passing between two interacting nodes.

\subsubsection{Attention-based Graph Filter} 
The attention-based graph filter is flexible and learnable, i.e., there is no constraint on $\mathcal{M}$ and the coefficients are data-driven. Given $\mathcal{M}$ and $\tilde{\mathbf{A}}^{i}$ of distance $i$ ($i\in\mathcal{M}$), the corresponding coefficient satisfies:
\begin{align} \label{eq5}
    \xi_i = \frac{\operatorname{exp}\left(\zeta_i\right)}{\sum_{j\in \mathcal{M}}\operatorname{exp}\left(\zeta_j\right)}, \ \ \forall i\in\mathcal{M},
\end{align}
which is the common softmax operation, and $\zeta_j$ ($j\in\mathcal{M}$) are free parameters that can be initialized as all zeros or appropriate priors and directly learned via backpropagation. Thus, the coefficients of $f_\xi$ can be interpreted as the attention scores assigned to neighborhoods of different hops. Note that although this attention mechanism is not node-adaptive (i.e., shared across graph nodes), this weight sharing fashion helps prevent overfitting and performs well on all the benchmark datasets.

\subsubsection{Diffusion-based Graph Filter}
In recent works~\cite{pagerank,gdc,scaling}, graph diffusion has demonstrated promising performance by propagating features using personalized PageRank (PPR) and heat kernel (HK). In this work, we evaluate the performance of PPR matrix as the fixed diffusion-based graph filter (HK is omitted due to comparable performance) in NCGNN:
\begin{align} \label{eq6}
    \bar{\mathbf{A}}= \alpha\left(\mathbf{I}_N-\left(1-\alpha\right)\tilde{\mathbf{A}}\right)^{-1}= \sum_{i=0}^{\infty}\alpha\left(1-\alpha\right)^{i}\tilde{\mathbf{A}}^{i},
\end{align}
where $\alpha \in (0,1)$ is termed teleport probability, $\mathcal{M}=\mathbb{N}$, and $\xi_i=\alpha(1-\alpha)^i$. When tackling large-scale graphs for which it is inefficient to compute matrix inversion, we use the truncated form as in Eq.~\eqref{eq7} by restricting the sum to a finite number $P$, similar to~\cite{pagerank}. We also apply the sparsification operation to the fixed diffusion-based graph filter to identify a subset of important graph nodes for feature aggregation.
\begin{align} \label{eq7}
    \bar{\mathbf{A}} \approx \sum_{i=0}^{P}\alpha\left(1-\alpha\right)^{i}\tilde{\mathbf{A}}^{i}.
\end{align}

\subsubsection{Discussion}
The designed graph filter in Eq.~\eqref{eq4} can be explained from both spatial and spectral aspects. In the spatial domain, each entry $\bar{a}_{ij}$ of $\bar{\mathbf{A}}$ represents the influence of the message passing from node $v_j$ to node $v_i$. In the spectral domain, the following proposition indicates the relationship between $\bar{\mathbf{A}}$ and the polynomial filter $g_{\theta}(\mathbf{L})$:
\begin{mypro}
Consider the graph $\bar{\mathcal{G}}$ with added self-loops and choose the graph Laplacian as $\mathbf{L}=\mathbf{I}_N-\tilde{\mathbf{A}}$. The graph filter defined by Eq.~\eqref{eq4} can be expressed as a $J$-order polynomial filter $g_{\theta}(\mathbf{L})=\sum_{j=0}^{J}\theta_j\mathbf{L}^{j}$ with $J=\max(\mathcal{M})$ in the spectral domain.
\begin{IEEEproof}
Rewrite $\bar{\mathbf{A}}$ as $f_\xi(\tilde{\mathbf{A}})=\sum_{i=0}^{I}\xi_i\tilde{\mathbf{A}}^{i}$ with $I=\max(\mathcal{M})$ and $\xi_i=0$ for $i\not\in\mathcal{M}$. The equivalence between $f_\xi(\tilde{\mathbf{A}})$ and $g_\theta(\mathbf{L})$, i.e., $\sum_{i=0}^{I}\xi_i\tilde{\mathbf{A}}^{i}=\sum_{j=0}^{J}\theta_j\mathbf{L}^{j}$, can be constructed via $\theta_j=\sum_{i=j}^{I}\binom{i}{j}\xi_i(-1)^{j}$ and $J=I$~\cite{gdc}.
\end{IEEEproof}
\end{mypro}

\subsection{Capsule Graph Layer}
After obtaining the subgraph nodes $\{v_j|\bar{a}_{ij}\neq0\}$ identified by the graph filter $\bar{\mathbf{A}}$ and a set of primary node capsules $\mathcal{S} = \{\mathbf{h}_{j}^{(k)} | \bar{a}_{ij}\neq0, k=1,\ldots,K\}$ for each target node $v_i \in \mathcal{V}$, this section elucidates how the designed capsule graph layer transforms node $v_i$ from $K$ primary node capsules to $C$ class node capsules $\mathcal{T}=\{\mathbf{u}_i^{(l)}|l=1,\ldots,C\}$ for classification with adaptive feature aggregation from $\mathcal{S}$ to $\mathcal{T}$. Note that although capsule graph layers can be further stacked to generate multiple high-level capsules in intermediate layers if necessary, we use one capsule graph layer in our study to directly produce class node capsules attributed to its efficiency, scalability, and remarkable performance. More discussions are provided in the Analysis Section~\ref{analysis}.

Denote with $f_c$ the dimension of class node capsules, $\mathbf{W}_{kl}\in\mathbb{R}^{f_c\times f_p}$ the learnable weights matrix between the $k$-th primary node capsules and the $l$-th class node capsules, and $\mathbf{b}_{l}\in\mathbb{R}^{f_c}$ the learnable bias for generating the $l$-th class node capsules. The feature generation rule for the $l$-th class node capsule $\mathbf{u}_{i}^{(l)}$ ($l=1,\ldots,C$) of node $v_i$ consists of Capsule Selection (\textbf{CS}) and Class Message Passing (\textbf{CMP}):
\begin{align} 
    \mathbf{p}_j^{(l)}&=\sum_{k=1}^{K}c_{jkl}\mathbf{W}_{kl}\mathbf{h}_{j}^{(k)}, \ \ \forall j\in\{j|\bar{a}_{ij}\neq0\} &(\textbf{CS}) \label{eq8} \\
    \mathbf{u}_i^{(l)}&=\sum\nolimits_{j\in\{j|\bar{a}_{ij}\neq0\}}\bar{a}_{ij}\mathbf{p}_j^{(l)}+\mathbf{b}_{l}  &(\textbf{CMP}) \label{eq9}
\end{align}
where $c_{jkl}$ is the coupling coefficient between $\mathbf{h}_j^{(k)}$ and $\mathbf{p}_j^{(l)}$, an intermediate capsule that represents the properties of class $l$ extracted from node $v_j$ and is used for generating the $l$-th class node capsules via CMP in Eq.~\eqref{eq9}. As per dynamic routing, $c_{jkl}$ ($k=1,\ldots,K$ and $l=1,\ldots,C$) are computed via an iterative neighborhood-routing-by-agreement procedure to adaptively select an appropriate subset of primary node capsules for aggregation from node $v_j$, which will be discussed later. $\bar{a}_{ij}$ is computed by the graph filter $\bar{\mathbf{A}}$, determining the influence of the message passing from node $v_j$ to node $v_i$. As can be seen in Eqs.~\eqref{eq8} and~\eqref{eq9}, $\mathbf{u}_i^{(l)}$ only aggregates the primary node capsules that are in agreement (i.e., larger $c_{jkl})$ to class $l$ from $\mathcal{S}$, while harmful noise is adaptively restrained (i.e., smaller $c_{jkl}$) to relieve the risk of feature over-mixing.

As in CapsNet~\cite{caps1}, we want the length of the output vector of each class node capsule $\mathbf{u}_{i}^{(l)}$ to indicate the presence of an instance of class $l$. Hence, we use the ``squashing'' activation function as in Eq.~\eqref{eq10} to ensure that the length of the output vector of $\mathbf{u}_{i}^{(l)}$ lies between zero and one, representing the probability that node $v_i$ belongs to class $l$. We also use the separate margin loss as in Eq.~\eqref{eq11} for each class node capsule.
\begin{align} 
\mathbf{v}_{i}^{(l)}&=\operatorname{squash}(\mathbf{u}_{i}^{(l)})=\frac{\|\mathbf{u}_{i}^{(l)}\|^2}{1+\|\mathbf{u}_{i}^{(l)}\|^2}\cdot \frac{\mathbf{u}_{i}^{(l)}}{\|\mathbf{u}_{i}^{(l)}\|}, \label{eq10} \\
\mathcal{L}_{i}^{(l)} &=T_{i}^{(l)}\cdot\max(0, m^{+}-\|\mathbf{v}_{i}^{(l)}\|)^{2} \nonumber\\
&\qquad+\lambda(1-T_{i}^{(l)})\cdot\max(0,\|\mathbf{v}_{i}^{(l)}\|-m^{-})^{2},\label{eq11}
\end{align}
where $\mathcal{L}_{i}^{(l)}$ is the loss for the $l$-th class node capsule of node $v_i$, $\mathbf{v}_{i}^{(l)}$ is the output vector of capsule $\mathbf{u}_{i}^{(l)}$, $T_{i}^{(l)}=1$ if node $v_{i}$ belongs to class $l$ and 0 otherwise, $m^{+}$ is a constant close to 1, $m^{-}$ is close to 0, and $\lambda$ is a parameter for preventing all class node capsules from shrinking to 0 at the beginning of training. The total loss is $\mathcal{L}=\frac{1}{|\mathcal{V}_l|}\sum_{v_i\in\mathcal{V}_l}\sum_{l=1}^{C}\mathcal{L}_{i}^{(l)}$.

\subsubsection*{Neighborhood-routing-by-agreement} 
While in CapsNet and CapsGNN, the coupling coefficients between a capsule and all the capsules in the layer above sum to $1$, which implies that each capsule gets sent to an appropriate parent in the next layer, our neighborhood-routing-by-agreement results from different motivations. In our NCGNN framework, the coupling coefficients between a class node capsule and the primary node capsules of each individual node in the subgraph sum to 1, i.e., $\sum_{k=1}^{K}c_{jkl}=1$. This means that for each subgraph node, some of its primary node capsules are in agreement, i.e., to be aggregated, while some are considered as noise to be restrained. The agreement is in the form of dot product of two vectors, and $c_{jkl}$ is iteratively adjusted according to Eqs.~\eqref{eq12} and~\eqref{eq13} for routing iterations $t=0,\ldots,T-1$, starting by initializing $c_{jkl}$ as $c_{jkl}^{(0)}=\operatorname{exp}(b_{jkl}^{(0)})/\sum_{m=1}^{K}\operatorname{exp}(b_{jml}^{(0)})$ with $b_{jkl}^{(0)}$ ($k=1,\ldots,K, l=1,\ldots,C$) of all zeros or set as some other appropriate priors:
\begin{align}
c_{jkl}^{(t)} &= \frac{\operatorname{exp}(b_{jkl}^{(t)})}{\sum_{m=1}^{K}\operatorname{exp}(b_{jml}^{(t)})}, \label{eq12} \\
b_{jkl}^{(t+1)} &= b_{jkl}^{(t)} + (\mathbf{v}_j^{(l)})^\top(\mathbf{W}_{kl}\mathbf{h}_{j}^{(k)}), \label{eq13}
\end{align}
where $\mathbf{v}_j^{(l)}$ is computed in parallel with $\mathbf{v}_i^{(l)}$. As can be seen, if the output of the class node capsule $\mathbf{v}_j^{(l)}$ has a large dot product with $\mathbf{W}_{kl}\mathbf{h}_{j}^{(k)}$, the coupling coefficient $c_{jkl}$ of node $v_j$ associated with $\mathbf{h}_{j}^{(k)}$ and class $l$ would be iteratively increased. In this way, for the generation of each class node capsule, the neighborhood-routing-by-agreement adaptively identifies a subset of capsules (with larger $c_{jkl}$) for aggregation from each subgraph node, which has the following two advantages: (i) Feature over-mixing of interacting nodes can be suppressed to some extent, as irrelevant messages from the subgraph are restrained, which helps relieve over-smoothing in larger receptive field and learn effectively over disassortative graphs; (ii) The selected primary node capsules that are in agreement with large coupling coefficients can be used to explain the model prediction as they are most influential for generating the corresponding class node capsules in label prediction. Detailed interpretability analysis is discussed in Section~\ref{interpret}.

The procedure of the capsule graph layer in node-wise fashion is summarized in Algorithm~\ref{alg}, which consists of only differentiable operations and can be optimized via gradient-based backpropagation.

\begin{algorithm}[t]
\caption{Capsule Graph Layer ($T$ routing iterations)}
\label{alg}
\textbf{Input}: Subgraph nodes $\{v_j|\bar{a}_{ij}\neq0\}$ identified by $\bar{\mathbf{A}}$ for the target node $v_i$, and a set of primary node capsules $\mathcal{S} = \{\mathbf{h}_{j}^{(k)} | \bar{a}_{ij}\neq0, k=1,\ldots,K\}$. \\
\textbf{Parameters}: $\mathbf{W}_{kl}, \mathbf{b}_l$, $k=1,\ldots,K, \ l=1,\ldots,C$.\\
\textbf{Output}: Output vectors of $C$ class node capsules $\{\mathbf{v}_i^{(l)}|l=1,\ldots,C\}$ for node $v_i$.
\begin{algorithmic}[1] 
\STATE For all nodes: $b_{jkl}^{(0)} \leftarrow 0 \text{ } \text{or other appropriate prior}$, \ $k=1,\ldots,K, \ l=1,\ldots,C$. \\
\FOR{routing iteration $t=0,\ldots,T-1$}
\STATE $c_{jkl}^{(t)} \leftarrow \operatorname{exp}(b_{jkl}^{(t)})/\sum_{m=1}^{K}\operatorname{exp}(b_{jml}^{(t)})$. 
\FOR{$l=1,\ldots, C$}
\STATE For all nodes: $\mathbf{p}_j^{(l)}\leftarrow \sum_{k=1}^{K}c_{jkl}^{(t)}\mathbf{W}_{kl}\mathbf{h}_{j}^{(k)}$, \\
\STATE $\mathbf{u}_i^{(l)}\leftarrow\sum_{j\in\{j|\bar{a}_{ij}\neq0\}}\bar{a}_{ij}\mathbf{p}_j^{(l)}+\mathbf{b}_{l}$, \\
\STATE $\mathbf{v}_i^{(l)}\leftarrow\operatorname{squash}(\mathbf{u}_i^{(l)})$ with Eq.~\eqref{eq10}, \\
\STATE $b_{jkl}^{(t+1)} \leftarrow b_{jkl}^{(t)} + (\mathbf{v}_j^{(l)})^\top(\mathbf{W}_{kl}\mathbf{h}_{j}^{(k)})$, $k=1,\ldots,K$. \quad \text{// $\mathbf{v}_j^{(l)}$ is computed in parallel with $\mathbf{v}_i^{(l)}$.}
\ENDFOR
\ENDFOR
\STATE \textbf{return} $\{\mathbf{v}_i^{(l)}|l=1,\ldots,C\}$.
\end{algorithmic}
\end{algorithm}

\subsection{Analysis} \label{analysis}
\subsubsection{Complexity Analysis} Let $N$ denote the number of graph nodes needed for computation, $L$ be the number of layers of GCN~\cite{gcn}, $T$ be the routing iterations of NCGNN, and $D$ be the dimension of node representations. For simplicity, we ignore the normalization step in Eq.~\eqref{eq3}, the activation function, the bias terms in Eqs.~\eqref{eq3} and~\eqref{eq9}, and the softmax function of Eq.~\eqref{eq12}. We also assume that $D=Kf_p=Cf_c$ is fixed for all layers. The time complexity of GCN is $\mathcal{O}(LND^{2}+L\|\tilde{\mathbf{A}}\|_0D)$, and that of NCGNN is about $\mathcal{O}((T+1)ND^{2}+T\|\bar{\mathbf{A}}\|_0D+2TNKD)$. Although our graph filter $\bar{\mathbf{A}}$ contains more higher-order neighborhood information than the first-order normalized adjacency matrix $\tilde{\mathbf{A}}$ utilized in GCN, $\bar{\mathbf{A}}$ is processed with $\epsilon$-threshold or top-$k$ sparsification, as illustrated above. Thus, the time complexity of our NCGNN framework is still comparable with GCN. Moreover, NCGNN shows scalability in a distributed manner~\cite{scaling} for processing large graphs, as it only has one capsule graph layer in our work to directly incorporate multi-hop neighborhood features, which is exempt from the exponential neighborhood expansion issue with the increasing message passing layers, i.e., to generate the class node capsules for each target node, we only need to load the features of its neighboring nodes identified by the sparsified graph filter $\bar{\mathbf{A}}$.

\subsubsection{Permutation Invariance}
Permutation invariance or equivariance is essential for designing aggregators in GNNs. In this section, we prove by induction that our proposed capsule graph layer described in Algorithm~\ref{alg} is permutation invariant to the order of graph nodes.
\begin{lemma}\label{lem1}
Consider the input set $Z=\{z_1,\ldots,z_M\}$, $z_m\in\mathcal{Z}$, i.e., the input domain is the power set $2^\mathcal{Z}$. 
 
If a function $f_1: 2^\mathcal{Z}\rightarrow\mathbb{R}^{n_1}$ operating on $Z$ can be decomposed in the form of $\rho\left(\sum_{z_m\in Z}\phi_2\left(f_2(Z),\phi_1(z_m)\right)\right)$ for suitable transformations $\phi_1$, $\phi_2$, and $\rho$ and another permutation invariant function $f_2: 2^\mathcal{Z}\rightarrow\mathbb{R}^{n_2}$ operating on $Z$, then $f_1$ is permutation invariant to the order of elements in $Z$.
\begin{IEEEproof}
Permutation invariance implies that the elements of a set have no particular order, and consequently, any function on the set must be constant to arbitrary ordering of its elements. Since $f_2$ is permutation invariant and addition is commutative, the function $\rho\left(\sum_{z_m\in Z}\phi_2\left(f_2(Z),\phi_1(z_m)\right)\right)$ satisfies permutation invariance.
\end{IEEEproof}
\end{lemma}
\begin{mypro}
Given the graph $\mathcal{G}$ and the graph filter $\bar{\mathbf{A}}$, the message passing scheme described in Algorithm~\ref{alg} is permutation invariant for arbitrary routing iteration.
\begin{IEEEproof}
Let us define $\mathbf{h}=\operatorname{PNC}(\mathbf{x})$ and $\mathbf{p}=\operatorname{CS}(\mathbf{c},\mathbf{h})$ according to Eqs.~\eqref{eq3} and \eqref{eq8}, respectively. Here, $\mathbf{x}$ is the given node features and $\mathbf{c}$ is the vectorized coupling coefficients. Note that the superscripts are omitted for simplicity. The input set is defined as $Z=\{\mathbf{x}_j|v_j\in\mathcal{V}\}$.

From Lemma~\ref{lem1}, $\mathbf{v}_i^{(l)}$ is permutation invariant to the ordering of elements in $Z$ at routing iteration $t=0$, as $\mathbf{v}_i^{(l)}=\rho\left(\sum_{\mathbf{x}_j\in Z}\phi_2\left(f_2(Z),\phi_1(\mathbf{x}_j)\right)\right)$ with $\phi_1(\mathbf{x}_j)=\operatorname{PNC}(\mathbf{x}_j)$, $f_2(Z)=\mathbf{c}^{(0)}$, $\phi_2(f_2(Z),\phi_1(\mathbf{x}_j))=\bar{a}_{ij}\cdot\operatorname{CS}(f_2(Z),\phi_1(\mathbf{x}_j))$, and $\rho\left(\square\right)=\operatorname{squash}\left(\square+\mathbf{b}_l\right)$. Here, $\mathbf{c}^{(0)}$ is initialized as constant values and thus $f_2$ is permutation invariant.

\emph{Induction Step}: Assume that for the $t$-th routing iteration, the message passing scheme of the capsule graph layer is permutation invariant. We consider the $(t+1)$-th routing iteration.
Since $\mathbf{v}_j^{(l)}$ ($v_j\in\mathcal{V}$) computed in the $t$-th routing iteration is permutation invariant, $b_{jkl}^{(t+1)}$ computed via Eq.~\eqref{eq13} are constant for any permutation. Thus, the coupling coefficients $\mathbf{c}^{(t+1)}$ are permutation invariant. Let $f_2(Z)=\mathbf{c}^{(t+1)}$, and consequently, $\mathbf{v}_i^{(l)}$ is permutation invariant at the end of the $(t+1)$-th routing iteration in the way similar to $t=0$. Therefore, the induction hypothesis is verified for the $(t+1)$-th routing iteration. As a result, we conclude that Algorithm~\ref{alg} is permutation invariant for any routing iteration.
\end{IEEEproof}
\end{mypro}

\subsubsection{Expressive Power}
We analyze the expressive power of our method by comparing with the naive sum aggregator.
\begin{mypro}
The message passing scheme of capsule graph layer is more powerful than the naive sum aggregator. That is the naive sum aggregator is a special case of our capsule graph layer.
\begin{IEEEproof}
Consider the following naive sum aggregator without nonlinear activation function:
\begin{align} \label{eq14}
\mathbf{h}_i^{\prime}=\sum\nolimits_{j\in\{j|\bar{a}_{ij}\neq0\}}\bar{a}_{ij}\mathbf{W}\mathbf{h}_j, \ \ \forall v_i \in \mathcal{V},
\end{align}
where $\mathbf{h}_i \in \mathbb{R}^{f_i}$ and $\mathbf{h}_i^{\prime} \in \mathbb{R}^{f_o}$ are input and output node features, respectively, and $\mathbf{W}\in\mathbb{R}^{{f_o}\times {f_i}}$ is the matrix of learnable weights for feature transformation.

Let us denote by $K$ and $C$ the numbers of each node's input capsules and output capsules, respectively. For arbitrary $v_j\in\mathcal{V}$, $k=1,\cdots,K$ and $l=1,\cdots,C$, we suppose $\mathbf{h}_j^{(k)}=\mathbf{h}_j$ for the $k$-th input capsule of $v_j$, $\mathbf{W}_{kl}=\mathbf{W}$, $\mathbf{b}_l$ are all constant zeros, and $b_{jkl}^{(0)}=0$. Thus, $c_{jkl}^{(0)}=1/K$. For the first routing iteration, the output capsules of each node can be generated by
\begin{align}
\mathbf{p}_j^{(l)}&=\sum_{k=1}^{K}c_{jkl}^{(0)}\mathbf{W}_{kl}\mathbf{h}_{j}^{(k)}=\mathbf{W}\mathbf{h}_{j}, \label{eq15} \\
\mathbf{u}_i^{(l)}&=\sum\nolimits_{j\in\{j|\bar{a}_{ij}\neq0\}}\bar{a}_{ij}\mathbf{p}_j^{(l)}=\mathbf{h}_i^{\prime}. \label{eq16}
\end{align}

Eq.~\eqref{eq16} implies that, for each node, its output capsules are all equal to its output node features shown in Eq.~\eqref{eq14}. Thus, $b_{jkl}^{(1)}$ of each node are equal according to Eq.~\eqref{eq13}, and consequently $c_{jkl}^{(1)}=1/K$. 

We can easily prove by induction that each node's output capsules are $C$ repetitions of its output node features for any routing iteration. Under this specific condition, the naive sum aggregator is a special case of our capsule graph layer.
\end{IEEEproof}
\end{mypro}

\section{Experiments} \label{sec4}
We evaluate NCGNN on the task of semisupervised node classification under various scenarios. Extensive experiments on twelve real-world graph datasets as well as synthetic graphs show that NCGNN outperforms a variety of state-of-the-art methods under both homophily and heterophily and well addresses the over-smoothing problem in GNNs. We also provide visualization results to demonstrate the inherent interpretability of our proposed message passing scheme.

\subsection{Datasets}
\subsubsection{Real-world Graphs}
Twelve real-world graph benchmarks are considered in the experiments, including six assortative graphs and six disassortative graphs. 

\begin{itemize}
    \item \textbf{Assortative Graphs}: (i) Cora~\cite{collective}, Cora-ML~\cite{automating,gauss}, Citeseer~\cite{collective}, and Pubmed~\cite{query} are four citation networks, where nodes represent documents, edges denote citations, node features are bag-of-words representations of corresponding papers, and node labels represent papers' academic topics; (ii) Amazon Photo and Amazon Computers~\cite{pitfall} are two segments of the Amazon co-purchase graph, where nodes are goods, edges indicate that two goods are frequently bought together, node features are bag-of-words encoded product reviews, and class labels are given by the product category. For each dataset, we use 20 labeled nodes per class for training, 500 nodes for validation, and the rest is set for test. For a fairer comparison, we select 5 random train/validation/test splits for each dataset and run 10 turns with different weight initialization seeds in each split, as suggested by~\cite{measuring}. Then we measure the average test accuracy of all the 50 runs’ results. We further take Amazon Computers as an example dataset to evaluate the model performance under different training set sizes.
    \item \textbf{Disassortative Graphs}: (i) Squirrel and Chameleon are two subgraphs of Wikipedia networks~\cite{wiki}, in which nodes represent web pages, edges indicate mutual links between pages, node features are bag-of-words representations corresponding to some informative nouns in the Wikipedia pages, and graph nodes are classified into five categories in terms of the number of the average monthly traffic of the page; (ii) Actor is an actor co-occurrence network extracted from the film-director-actor-writer network~\cite{actor}, where nodes denote actors, edges indicate co-occurrence on the same Wikipedia page, node features are bag-of-words representations of some keywords in the Wikipedia pages, and each node is labeled with one of the five categories according to the words of actor’s Wikipedia; (iii) Texas, Cornell and Wisconsin are three subgraphs from the CMU WebKB dataset, in which nodes represent web pages, edges are hyperlinks, node features are bag-of-words vectors representing corresponding web pages, and web pages are classified into the five categories: course, faculty, student, project, staff. For all the dissartative graphs, we utilize the preprocessed version in~\cite{beyond}. We use the same 10 random train/validation/test splits provided by~\cite{geomgcn,beyond} and measure the model performance by the average test accuracy over the 10 splits with 10 random weight initializations for each split.
\end{itemize}

Table~\ref{tab1} summarizes the statistics of these real-world benchmark datasets.

\begin{table*}[!t]
\renewcommand{\baselinestretch}{1.0}
\renewcommand{\arraystretch}{1.0}
\setlength{\tabcolsep}{3.8pt}
\centering
\caption{Statistics of real-world graphs.}
\begin{tabular}{l|cccccc|cccccc}
\toprule
&  \multicolumn{6}{c|}{\textbf{Assortative graphs}}  &  \multicolumn{6}{c}{\textbf{Disassortative graphs}} \\
\textbf{Datasets} & Cora & Citeseer & Pubmed & Cora-ML & Amazon Photo & Amazon Computers & Squirrel & Chameleon & Actor & Texas & Cornell & Wisconsin \\
\midrule
$H(\mathcal{G})$ & 0.81 & 0.74 & 0.80 & 0.79 & 0.83 & 0.78 & 0.22 & 0.23 & 0.22 & 0.06 & 0.30 & 0.18 \\
\# of Nodes & 2708 & 3312 & 19717 & 2995 & 7535 & 13471 & 5201 & 2277 & 7600 & 183 & 183 & 251 \\
\# of Edges & 5278 & 4536 & 44324 & 8158 & 119081 & 245861 & 198353 & 31371 & 26659 & 279 & 277 & 450 \\
\# of Features & 1433 & 3703 & 500 & 2879 & 745 & 767 & 2089 & 2325 & 932 & 1703 & 1703 & 1703\\
\# of Classes  & 7 & 6 & 3 & 7 & 8 & 10 & 5 & 5 & 5 & 5 & 5 & 5 \\
\bottomrule
\end{tabular}
\label{tab1}
\end{table*}

\subsubsection{Synthetic Graphs}
We further validate our proposed NCGNN on the synthetic graphs termed Syn-Cora~\cite{beyond}, where the homophily/heterophily level can be controlled. The graph generation process is described in~\cite{beyond}. Specifically, we generate 9 graphs, each with a different homophily ratio ranging from 0.1 to 0.9 at 0.1 intervals. An overview of dataset statistics is demonstrated in Table~\ref{tab2}. Following~\cite{beyond}, we randomly split nodes into $25\%$, $25\%$, $50\%$ for training, validation, and test. We also measure the average test accuracy over 10 random weight initializations for each graph.

\begin{table}[!t]
\renewcommand{\baselinestretch}{1.0}
\renewcommand{\arraystretch}{1.0}
\setlength{\tabcolsep}{3.5pt}
\centering
\caption{Statistics of the dataset of synthetic graphs, \emph{i.e.}, Syn-Cora.}    \label{tab2}
\begin{tabular}{ccccc}
\toprule
\textbf{\# of Nodes} & \textbf{\# of Edges} & \textbf{\# of Features} & \textbf{\# of Classes} & \textbf{\# of Graphs}\\
\midrule
1490 & 2965 to 2968 & 1433 & 5 & 9\\
\bottomrule
\end{tabular}
\end{table}

\subsection{Experimental Set-up}
\subsubsection{Baselines}
For node classification on assortative graphs, we compare with the following baselines. (i) Three representative GNNs: GCN~\cite{gcn}, GAT~\cite{gat}, and SGC~\cite{sgc}. (ii) GNN with graph diffusion (i.e., PPR) based message passing: APPNP~\cite{pagerank} (PPNP~\cite{pagerank} is omitted due to comparable performance with APPNP). (iii) Four recently proposed GNNs for tackling over-smoothing: JK-Net~\cite{jknet}, DAGNN~\cite{dagnn}, GCNII~\cite{gcnii}, and Scattering GCN~\cite{scattering}. (iv) Some other tricks with GCN as the backbone for relieving the over-smoothing issue: AdaEdge \cite{measuring}, DropEdge \cite{dropedge} and BBGDC \cite{bbgdc}. For learning over disassortative graphs, we consider the following baselines. (i) The simplest deep learning model that ignores the graph structure: MLP. (ii) Two popular GNNs: GCN and GAT. (iii) Two strong baselines in solving semisupervised node classification for assortative graphs: DAGNN and GCNII. (iv) Geom-GCN with different embedding methods~\cite{geomgcn}: Geom-GCN-I, Geom-GCN-P, and Geom-GCN-S. (v) Two variants of $\text{H}_{\text{2}}\text{GCN}$~\cite{beyond} which achieve the state-of-the-art performance under heterophily: $\text{H}_{\text{2}}\text{GCN-1}$ (one round for neighborhood aggregation) and $\text{H}_{\text{2}}\text{GCN-2}$ (two rounds for neighborhood aggregation). Note that DisenGCN~\cite{disen} leverages a message passing scheme similar to the multi-head mechanism adopted in GAT, as discussed in Section~\ref{message}. Since the source code of DisenGCN is not publicly available, and its reported performance is inferior to GCNII and hard to be reproduced, we do not include it as a baseline in our evaluations. 

\subsubsection{Model Configurations}
The dimension of class node capsules ($f_c$) is set to 16. Dropout of $p=0.9$ is applied to the primary node capsules. We use Adam~\cite{adam} optimizer to minimize the error function with a learning rate of $1e-3$. Training epochs are manually set for each dataset. $\ell_2$ regularization on the learnable weights is employed with weight decay of $1e-3$ for the six real-world disassortative graphs, and $5e-3$ for the others. $\mathcal{M}=\{1,\ldots,\operatorname{max}(\mathcal{M})\}$ or $\mathcal{M}=\{0,\ldots,\operatorname{max}(\mathcal{M})\}$ is employed for the attention-based graph filter and we choose the setting that achieves better performance on the validation set for each dataset. The teleport probability in Eq.~\eqref{eq6} is set to $\alpha \in \{0.05, 0.1\}$ for the diffusion-based graph filter. Graph filters are sparsified by top-$k$ sparsification with $k \in \{128, 256\}$ or $\epsilon$-threshold sparsification with $\epsilon=0.0001$. $\lambda$ is set to 0.5. The other hyperparameters are tuned in the following search space: $K\in\{4,6,8,10,12\}$, $f_p\in\{32,64,96,128\}$, $T\in\{2,3,\ldots,8\}$, $m^{-}\in\{0.05,0.1,0.15,0.2,0.25,0.3\}$, and $m^{+}\in\{0.7,0.75,0.8,0.85,0.9,0.95\}$.

\subsection{semisupervised Node Classification} 
\subsubsection{Citation Networks}
We compare NCGNN with the baselines under different sizes of receptive field on the four citation networks. We only consider the maximum hop of neighborhoods of 5 for NCGNN-A, i.e., $\mathcal{M}=\{1,\ldots,5\}$, with which the size of interacting neighborhoods is large enough to aggregate appropriate information and the potential risk of suffering from over-smoothing is high. Note that for most GNN models that only aggregate one-hop neighborhood information in a message passing layer (e.g., GCN), we stack multiple layers to enlarge the receptive field, while some other methods directly aggregate high-order neighborhood features (e.g., SGC). We also report the best performance for each baseline with the hyperparameter setting used in their corresponding papers. Table~\ref{tab3} summarizes the average test accuracy and the best accuracy in each column is highlighted in bold. 

\begin{table*}[t]
\centering
\caption{Average classification accuracy (\%) on the four citation networks, \emph{i.e.}, Cora, Citeseer, Pubmed, and Cora-ML.}
\begin{threeparttable}
\begin{tabular}{l|cccccc|cccccc}
\toprule
\textbf{Datasets} & \multicolumn{6}{c|}{Cora}  & \multicolumn{6}{c}{Citeseer} \\
\midrule
\textbf{\# of Hop} & 1 & 2 & 3 & 4 & 5 & best & 1 & 2 & 3 & 4 & 5 & best \\
\midrule
GCN~\cite{gcn} & 75.72 & 80.90 & 81.03 & 79.74 & 77.20 & 81.03 (3) & 66.66 & 68.58 & 66.96 & 64.84 & 59.30 & 68.58 (2) \\
GAT~\cite{gat} & 76.57 & 81.28 & 79.54 & 78.70 & 74.76 & 81.28 (2) & 68.39 & 68.79 & 66.19 & 62.85 & 58.32 & 68.79 (2) \\
SGC~\cite{sgc} & 75.72 & 80.13 & 80.88 & 81.16 & 80.98 & 81.16 (4) & 66.66 & 68.26 & 68.62 & 68.84 & 69.03 & 69.03 (5) \\
APPNP~\cite{pagerank}  & - & - & - & - & - & 82.08 (10)  & - & - & - & - & - & 68.99 (10) \\
JK-Net~\cite{jknet} & - & 80.13 & 80.34 & 79.52 & 77.68 & 80.34 (3) & - & 67.10 & 65.82 & 64.35 & 61.90 & 67.10 (2) \\
DAGNN~\cite{dagnn} & 72.93 & 77.72 & 79.91 & 81.19 & 81.91 & 83.25 (10) & 66.02 & 67.65 & 68.44 & 68.75 & 70.07 & 70.15 (10) \\
GCNII~\cite{gcnii} & 77.16 & 80.30 & 80.71 & 81.31 & 81.57 & 83.28 (64) & 65.97 & 67.01 & 66.85 & 66.84 & 67.15 & 69.48 (32) \\
Scattering GCN~\cite{scattering}  & - & - & - & - & - & 81.04 (17)  & - & - & - & - & - & 65.60 (65) \\
AdaEdge~\cite{measuring} & 75.62 & 80.79 & 79.73 & 78.33 & 76.27 & 80.79 (2) & 66.61 & 68.64 & 66.52 & 64.13 & 59.72 & 68.64 (2) \\
DropEdge~\cite{dropedge} & 75.58 & 80.56 & 80.71 & 79.94 & 77.31 & 80.71 (3) & 66.52 & 68.54 & 67.60 & 65.33 & 61.27 & 68.54 (2) \\
BBGDC~\cite{bbgdc} & - & 80.89 & 81.19 & 79.94 & 75.94 & 81.19 (3) & - & \textbf{69.87} & 68.11 & 64.48 & 63.00 & 69.87 (2) \\
\midrule
\textbf{NCGNN-D}  & - & - & - & - & - & \textbf{83.64 ($\infty$)}  & - & - & - & - & - & \textbf{71.20 ($\infty$)} \\
\textbf{NCGNN-A} & \textbf{79.69} & \textbf{81.81} & \textbf{82.21} & \textbf{83.18} & \textbf{83.62} & 83.62 (5) & \textbf{68.66} & 69.84 & \textbf{70.32} & \textbf{70.63} & \textbf{70.96} & 70.96 (5) \\
\bottomrule
\toprule
\textbf{Datasets} & \multicolumn{6}{c|}{Pubmed}  & \multicolumn{6}{c}{Cora-ML} \\
\midrule
\textbf{\# of Hop} & 1 & 2 & 3 & 4 & 5 & best & 1 & 2 & 3 & 4 & 5 & best \\
\midrule
GCN~\cite{gcn} & 74.65 & 77.44 & 75.90 & 75.32 & 73.55 & 77.44 (2) & 79.87 & \textbf{85.26} & 83.19 & 81.28 & 75.85 & 85.26 (2) \\
GAT~\cite{gat} & 74.51 & 77.62 & 75.08 & 74.19 & 65.39 & 77.62 (2) & 79.81 & 83.81 & 82.56 & 80.18 & 77.32 & 83.81 (2) \\
SGC~\cite{sgc} & 74.65 & 77.14 & 77.11 & 77.04 & 76.11 & 77.14 (2) & 79.87 & 85.01 & 84.25 & 83.57 & 82.93 & 85.01 (2) \\
APPNP~\cite{pagerank} & - & - & - & - & - & 78.30 (10) & - & - & - & - & - & 84.68 (10) \\
JK-Net~\cite{jknet} & - & 76.20 & 75.64 & 74.69 & 73.07 & 76.20 (2) & - & 82.94 & 81.43 & 81.06 & 79.40 & 82.94 (2) \\
DAGNN~\cite{dagnn} & 74.53 & 76.73 & 77.77 & 78.67 & 78.77 & 79.02 (20) & 79.03 & 82.37 & 82.70 & 84.19 & 84.64 & 85.95 (20) \\
GCNII~\cite{gcnii} & 75.00 & 78.44 & 78.48 & 78.49 & 78.37 & 78.94 (16) & 81.29 & 83.12 & 83.16 & 83.65 & 83.68 & 84.68 (64) \\
Scattering GCN~\cite{scattering}  & - & - & - & - & - & 75.97 (5) & - & - & - & - & - & 85.13 (17) \\
AdaEdge~\cite{measuring} & 74.67 & 77.16 & 73.53 & 71.00 & 68.82 & 77.16 (2) & 79.83 & 85.10 & 82.31 & 79.87 & 74.42 & 85.10 (2) \\
DropEdge~\cite{dropedge} & 74.21 & 77.47 & 76.87 & 75.86 & 74.56 & 77.47 (2) & 78.95 & 84.69 & 82.92 & 81.06 & 77.84 & 84.69 (2) \\
BBGDC~\cite{bbgdc} & - & OOM & OOM & OOM & OOM & OOM & - & 84.67 & 83.64 & 82.80 & 82.08 & 84.67 (2) \\
\midrule
\textbf{NCGNN-D}  & - & - & - & - & \textbf{78.94} & 78.94 (5)  & - & - & - & - & - & \textbf{86.14 ($\infty$)} \\
\textbf{NCGNN-A} & \textbf{76.45} & \textbf{78.56} & \textbf{79.20} & \textbf{79.06} & 78.93 & \textbf{79.20 (3)} & \textbf{83.32} & 85.23 & \textbf{85.71} & \textbf{85.82} & \textbf{85.90} & 85.90 (5) \\
\bottomrule
\end{tabular}
\begin{tablenotes}
\footnotesize
\item $^*$ ``-D'' denotes diffusion-based graph filter. ``-A'' denotes attention-based graph filter. ``-'' means that the corresponding setting is not applicable. OOM denotes out-of-memory error. The number in parentheses indicates the best-performing size of receptive field. Note that in our NCGNN, we compute the fully personalized PageRank matrix in Eq.~\eqref{eq6} for Cora, Citeseer, and Cora-ML, while use the truncated form in Eq.~\eqref{eq7} with $P=5$ for Pubmed.
\end{tablenotes}
\end{threeparttable}
\label{tab3}
\end{table*}

Overall, our NCGNN is able to achieve competitive performance on these four assortative graphs. Specifically, we make the following observations. (i) Compared with APPNP that also employs PPR to directly aggregate multi-hop neighborhood information, the classification accuracy improvement of NCGNN-D is significant. This fact verifies that the designed capsule graph layer indeed brings in more learning capacity than the naive sum aggregator to help learn effective node representations. (ii) AdaEdge, DropEdge, and BBGDC can help relieve the over-smoothing issue to some extent, but still suffer from drastic performance degradation when stacking more GCN layers to capture long-range dependencies. By contrast, NCGNN-A can achieve better classification accuracy with larger receptive field. (iii) DAGNN and GCNII are two strong baselines in solving semisupervised node classification for assortative graphs. Compared with these two methods, NCGNN achieves the state-of-the-art performance for all datasets.

\subsubsection{Amazon Co-purchase Graphs}

\begin{figure*}[t]
\centerline{\includegraphics[scale=0.16]{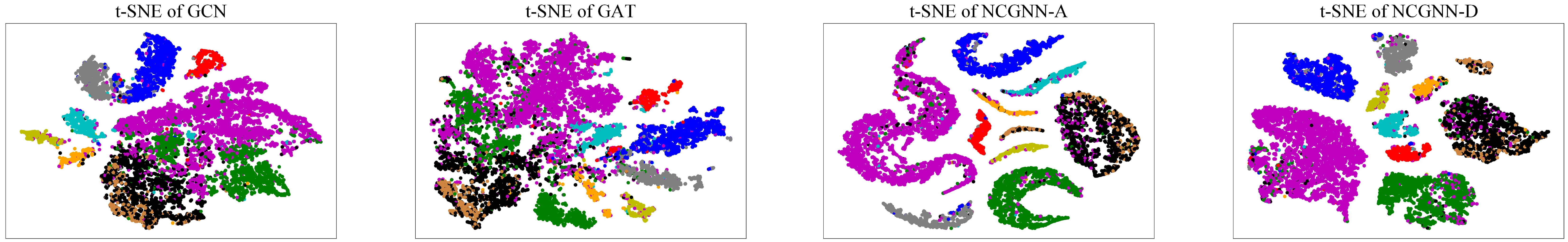}}
\caption{t-SNE plots of node representations obtained by GCN, GAT, and NCGNN on Amazon Computers, respectively. Different colors denote different classes.}
\label{fig3}
\end{figure*}

For Amazon co-purchase graphs, we just consider 2-hop neighbors for NCGNN-A as its size is large enough for aggregating useful information (the number of 2-hop neighbors of Amazon Photo is averagely 815 and that of Amazon Computers is about 1887). The average test accuracy is reported in Table~\ref{tab4}. The best accuracy in each column is highlighted in bold. It shows that NCGNN achieves the best performance on both datasets and outperforms the baselines by large margins on Amazon Computers. We argue that the performance boost is gained from the dynamic routing between node-level capsules. The reason is that previous message passing schemes would inevitably aggregate excessive noise in Amazon Computers where the average receptive field size is very large, while NCGNN can effectively restrain noise from the large neighborhoods to prevent feature over-mixing. We also plot t-SNE~\cite{tsne} of the node representations (hidden features for GCN and GAT, and the output vector of each node's class node capsule with the maximum length for NCGNN) learned on Amazon Computers in Fig.~\ref{fig3}. As can be seen, node embeddings generated by NCGNN are much more separable across different classes (colors), which also verifies that dynamic routing can indeed aggregate appropriate features and restrain harmful noise to relieve the over-smoothing issue.

\begin{table}[!t]
\renewcommand{\baselinestretch}{1.0}
\renewcommand{\arraystretch}{1.0}
\centering
\caption{Average test accuracy (\%) on the Amazon Co-purchase graphs}\label{tab4}
\begin{threeparttable}
\begin{tabular}{lcc}
\toprule
\textbf{Datasets} & Amazon Photo & Amazon Computers  \\
\midrule
GCN~\cite{gcn} & 92.21 & 82.45 \\
GAT~\cite{gat} & 89.36 & 82.80 \\
SGC~\cite{sgc} & 90.49 & 83.38 \\
APPNP~\cite{pagerank} & 91.96 & 81.39 \\
JK-Net~\cite{jknet} & 91.76 & 79.93 \\
DAGNN~\cite{dagnn} & 92.69 & 82.78 \\
GCNII~\cite{gcnii} & 89.88  & 81.13 \\
Scattering GCN~\cite{scattering} & 81.51 & 68.69 \\
AdaEdge~\cite{measuring} & 92.22 & 82.18 \\
DropEdge~\cite{dropedge} & 92.21 & 82.30 \\
BBGDC~\cite{bbgdc} & 90.59  & OOM \\
\midrule
\textbf{NCGNN-D} & 92.86 & 88.11 \\
\textbf{NCGNN-A} & \textbf{93.08} & \textbf{88.12}\\
\bottomrule
\end{tabular}
\begin{tablenotes}
\footnotesize
\item $^*$ The receptive field sizes are tuned and set as 10, 5, and 17 for APPNP, DAGNN and Scattering GCN, respectively, and 2 for the other baselines. OOM denotes out-of-memory error.
\end{tablenotes}
\end{threeparttable}
\end{table}

\begin{table}[!t]
\renewcommand{\baselinestretch}{1.0}
\renewcommand{\arraystretch}{1.0}     
\setlength{\tabcolsep}{2.2pt}
\centering
\caption{Average test accuracy (\%) on disassortative graphs.}\label{tab5}
\begin{threeparttable}
\begin{tabular}{lcccccc}
\toprule
\textbf{Datasets} & Squirrel & Chameleon & Actor & Texas & Cornell & Wisconsin \\
\midrule
MLP & 29.68 & 46.36 & 35.76 & 81.89 & 81.08 & 85.29 \\
GCN~\cite{gcn} & 36.89 & 59.82 & 30.26 & 59.46 & 57.03 & 59.80 \\
GAT~\cite{gat} & 30.62 & 54.69 & 26.28 & 58.38 & 58.92 & 55.29 \\
DAGNN~\cite{dagnn} & 35.57 & 52.40 & 33.68 & 59.35 & 59.89 & 55.84\\
GCNII~\cite{gcnii} & 35.55 & 55.32 & 33.85 & 75.46 & 78.11 & 78.04 \\
Geom-GCN-I~\cite{geomgcn} & 33.32 & 60.31 & 28.65 & 57.58 & 56.76 & 58.24 \\
Geom-GCN-P~\cite{geomgcn} & 38.14 & 60.90 & 31.28 & 67.57 & 60.81 & 64.12 \\
Geom-GCN-S~\cite{geomgcn} & 36.24 & 59.96 & 29.83 & 59.73 & 55.68 & 56.67 \\
$\text{H}_{\text{2}}\text{GCN-1}$~\cite{beyond} & 36.42 & 57.11 & 35.86 & \textbf{84.86} & 82.16 & 86.67 \\
$\text{H}_{\text{2}}\text{GCN-2}$~\cite{beyond} & 37.90 & 59.39 & 35.62  & 82.16 & 82.16 & 85.88 \\
\midrule
\textbf{NCGNN-A} & \textbf{48.80} & \textbf{63.97} & \textbf{36.84} & 84.43 & \textbf{83.79} & \textbf{86.90}\\
\bottomrule
\end{tabular}
\begin{tablenotes}
\footnotesize
\item $^*$ The results of DAGNN, GCNII, and NCGNN-A are obtained by our implementations, while the others are from~\cite{beyond,geomgcn}.
\end{tablenotes}
\end{threeparttable}
\end{table}

\begin{table*}[t]
\renewcommand{\baselinestretch}{1.0}
\renewcommand{\arraystretch}{1.0}
\setlength{\tabcolsep}{13pt}
\centering
\scriptsize
\caption{Average test accuracy (\%) on the Syn-Cora datasets}\label{tab6}
\begin{tabular}{lccccccccc}
\toprule
\textbf{Homophily Ratio} $H(\mathcal{G})$ & 0.1 & 0.2 & 0.3 & 0.4 & 0.5 & 0.6 & 0.7 & 0.8 & 0.9 \\
\midrule
MLP & 71.60 & 71.97 & 71.01 & 71.92 & 73.18 & 71.14 & 69.61 & 71.30 & 69.34 \\
GCN~\cite{gcn} & 32.83 & 38.95 & 49.77 & 55.49 & 65.13 & 77.29 & 83.60 & 90.18 & 96.19 \\
GAT~\cite{gat} & 30.71 & 36.46 & 47.27 & 53.91 & 64.56 & 75.17 & 81.96 & 89.37 & 95.27 \\
DAGNN~\cite{dagnn} & 48.30 & 55.25 & 63.54 & 66.28 & 75.01 & 83.09 & 87.09 & 91.97 & 96.97 \\
GCNII~\cite{gcnii} & 57.16 & 62.47 & 70.20 & 71.54 & 78.07 & 84.99 & 88.83 & \textbf{93.67} & 97.26 \\
$\text{H}_{\text{2}}\text{GCN-1}$~\cite{beyond} & 75.06 & 73.88 & 73.77 & 75.68 & 80.08 & 83.28 & 87.41 & 91.97 & 96.48 \\
$\text{H}_{\text{2}}\text{GCN-2}$~\cite{beyond} & 74.21 & 72.56 & 73.77 & 74.66 & 80.08 & 84.59 & 86.82 & 92.32 & 95.94 \\
\midrule
\textbf{NCGNN-A} & \textbf{75.44} & \textbf{74.42} & \textbf{74.87} & \textbf{77.39} & \textbf{81.48} & \textbf{85.18} & \textbf{90.74} & 93.61 & \textbf{97.42} \\
\bottomrule
\end{tabular}
\end{table*}

\begin{table}[t]
\renewcommand{\baselinestretch}{1.0}
\renewcommand{\arraystretch}{1.0}    
\centering
\scriptsize
\caption{Average test accuracy (\%) on the Amazon Computers dataset with varying training set sizes.}\label{tab7}
\begin{tabular}{lcccc}
\toprule
\textbf{\# of Labels per Class} & 5 & 10 & 20 & 30 \\
\midrule
GCN~\cite{gcn} & 73.21 & 79.73 & 82.45 & 83.92 \\
GAT~\cite{gat} & 73.86 & 76.87 & 82.80 & 83.25 \\
SGC~\cite{sgc} & 73.79 & 80.47 & 83.38 & 84.19 \\
APPNP~\cite{pagerank} & 71.28 & 77.82 & 81.39 & 82.09 \\
JK-Net~\cite{jknet} & 68.67 & 76.50 & 79.93 & 80.37 \\
DAGNN~\cite{dagnn} & 72.85 & 78.42 & 82.78 & 84.49 \\
GCNII~\cite{gcnii} & 65.88 & 73.21 & 81.13 & 82.73\\
Scattering GCN~\cite{scattering} & 50.79 & 65.44 & 68.69 & 69.11 \\
AdaEdge~\cite{measuring} & 73.09 & 79.68 & 82.18 & 83.79 \\
DropEdge~\cite{dropedge} & 73.36 & 80.03 & 82.30 & 83.98 \\
\midrule
\textbf{NCGNN-D} & 83.29 & \textbf{86.44} & 88.11 & \textbf{88.36} \\
\textbf{NCGNN-A} & \textbf{83.95} & 86.34 & \textbf{88.12} & 88.33 \\
\bottomrule
\end{tabular}
\end{table}

\subsubsection{Disassortative Graphs}
We now evaluate the performance of our method and existing baselines on six real-world disassortative graphs that exhibit low homophily ratio compared with citation netowrks and Amazon co-purchase graphs. We only consider NCGNN-A here since PPR is commonly based on the assumption of homophily~\cite{gdc}. We consider the receptive field as $\operatorname{max}(\mathcal{M})=3$. Experimental results are summarized in Table~\ref{tab5} and the best accuracy in each column is highlighted in bold. From the table, we can find the following facts. (i) The two popular GNN models (i.e., GCN and GAT) and the two strong baselines for learning over assortative graphs (i.e., DAGNN and GCNII) do not perform well on these disassortative graphs and are even outperformed than the simplest deep learning model (i.e., MLP), while our NCGNN-A achieves significant performance gains compared with all these methods. (ii) Compared with Geom-GCN and $\text{H}_{\text{2}}\text{GCN}$, NCGNN-A achieves the state-of-the-art performance on five disassortative graphs. Although NCGNN-A is outperformed by $\text{H}_{\text{2}}\text{GCN-1}$ on the Texas dataset, the number of misclassified nodes on the test set is averagely only one more than $\text{H}_{\text{2}}\text{GCN-1}$. We think the reason is that the Texas graph is too small (183 nodes) to demonstrate the superiority of our method.

\subsubsection{Synthetic Graphs}
Following~\cite{beyond}, we also use synthetic graphs to further validate the performance of our method. Table~\ref{tab6} shows the average test accuracy over 10 random weight initializations for each synthetic graph. We evaluate the following baselines: MLP, GCN, GAT, DAGNN, GCNII, and $\text{H}_{\text{2}}\text{GCN}$. The best accuracy in each column is highlighted in bold. As can be seen, NCGNN-A achieves the best performance with homophily ratio ranging from 0.1 to 0.7, and almost ties with GCNII under the strong homophily setting (i.e., $H(\mathcal{G})=0.8$ and $0.9$). The performance of GCN, GAT, DAGNN, and GCNII is very poor under the heterophily setting and consistently increases with $H(\mathcal{G})$ varying from 0.1 to 0.9. They are even outperformed by MLP that ignores the graph structure when $H(\mathcal{G})<0.5$, verifying that their message passing schemes cannot aggregate desirable features from the heterophily-dominant neighborhoods. $\text{H}_{\text{2}}\text{GCN}$ also demonstrates competitive performance under both heterophily and homophily compared with other baselines, but is outperformed by NCGNN-A in all cases.

\begin{figure}[t]
\centerline{\includegraphics[scale=0.215]{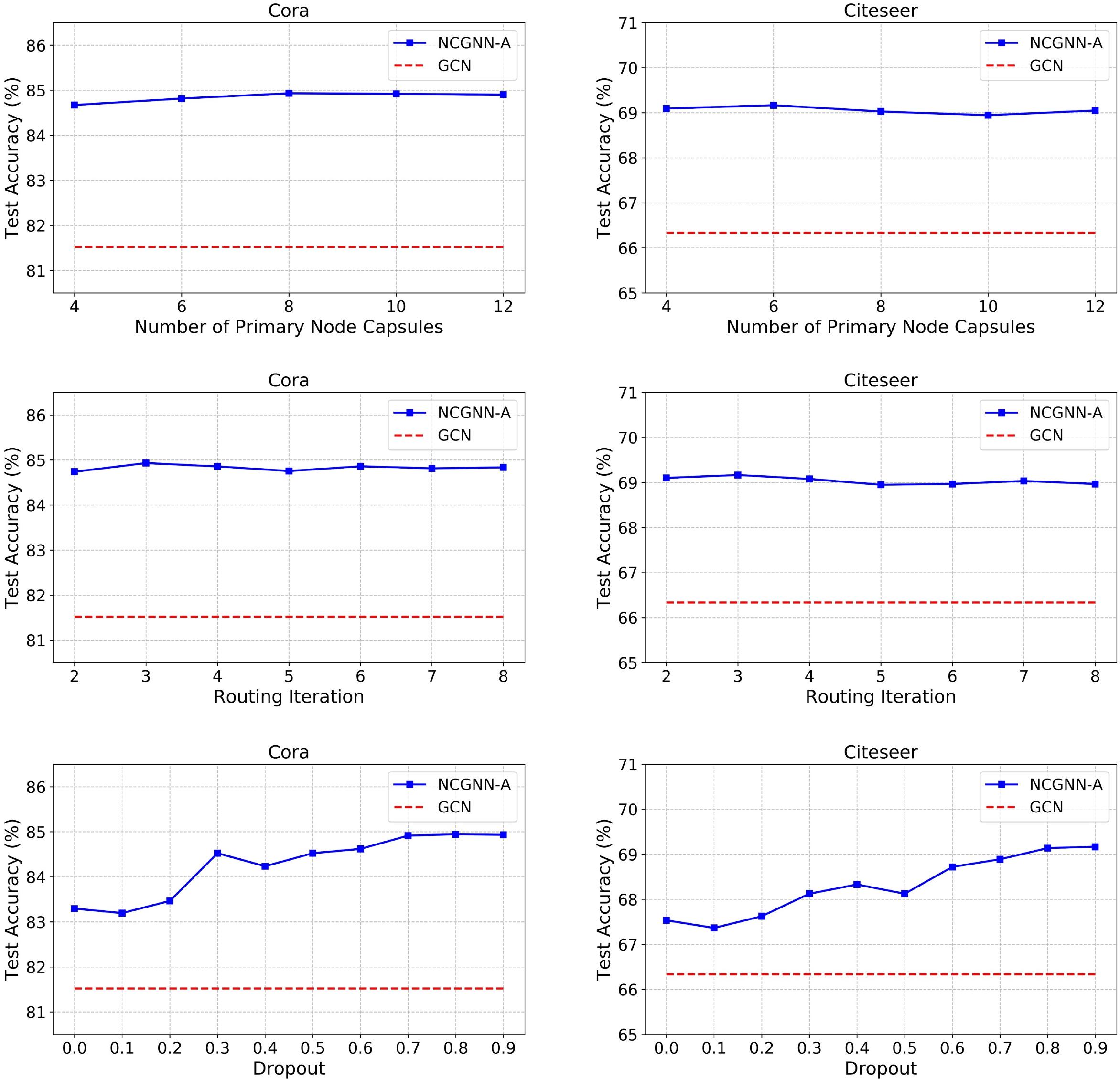}}
\caption{Hyperparameter sensitivity analysis of NCGNN on two citation networks, i.e., Cora and Citeseer.\label{fig4}}
\end{figure}

\subsection{Analysis}
\subsubsection{Effect of Training Set Size}
In semisupervised learning, the number of training samples is usually limited. In this section, we take Amazon Computers as an example dataset to evaluate the model performance with varying training set sizes (i.e., 5, 10, 20, and 30 labels per class). Experimental results are demonstrated in Table~\ref{tab7}. The best accuracy in each column is highlighted in bold. We can observe that NCGNN consistently achieves the best performance in all cases. Notably, when the training samples are severely scarce (i.e., 5 labels per class), the superiority of NCGNN becomes more pronounced. Moreover, compared with the two strong baselines (i.e., DAGNN and GCNII) that perform well on the four citation networks, NCGNN demonstrates remarkable improvement, which implies that NCGNN is more robust across different types of graph datasets. 

\subsubsection{Hyperparameter Sensitivity Analysis}
We investigate the impacts of three important hyperparameters in NCGNN, i.e., the number of primary node capsules ($K$), the routing iterations ($T$), and the dropout rate ($p$). Fig.~\ref{fig4} demonstrates the influence of these hyperparameters on the classification performance of NCGNN-A for Cora and Citeseer with a fixed dataset split. Note that NCGNN-D shows similar performance to NCGNN-A. The results indicate that NCGNN is not sensitive to $K$ and $T$. For $K$ ranging from 4 to 12 and $T$ ranging from 2 to 8, NCGNN consistently outperforms GCN by large margins. Note that more routing iterations should lead to better interpretability. Moreover, we find that our model needs a large dropout rate to avoid over-fitting, which is common in the case of complex models and scarce training data of semisupervised learning.

\subsubsection{Interpretability Analysis} \label{interpret}
The node-level capsule-based message passing is inherently interpretable by adaptively identifying a subset of input node capsules that are most significant to the model's prediction from the extracted subgraph. Note that the interpertability of primary node capsules has been addressed in existing methods
such as~\cite{inter_caps}. Therefore, we elaborate the inherent interpretability within the message passing scheme.

\begin{figure}[t]
\centerline{\includegraphics[scale=0.2]{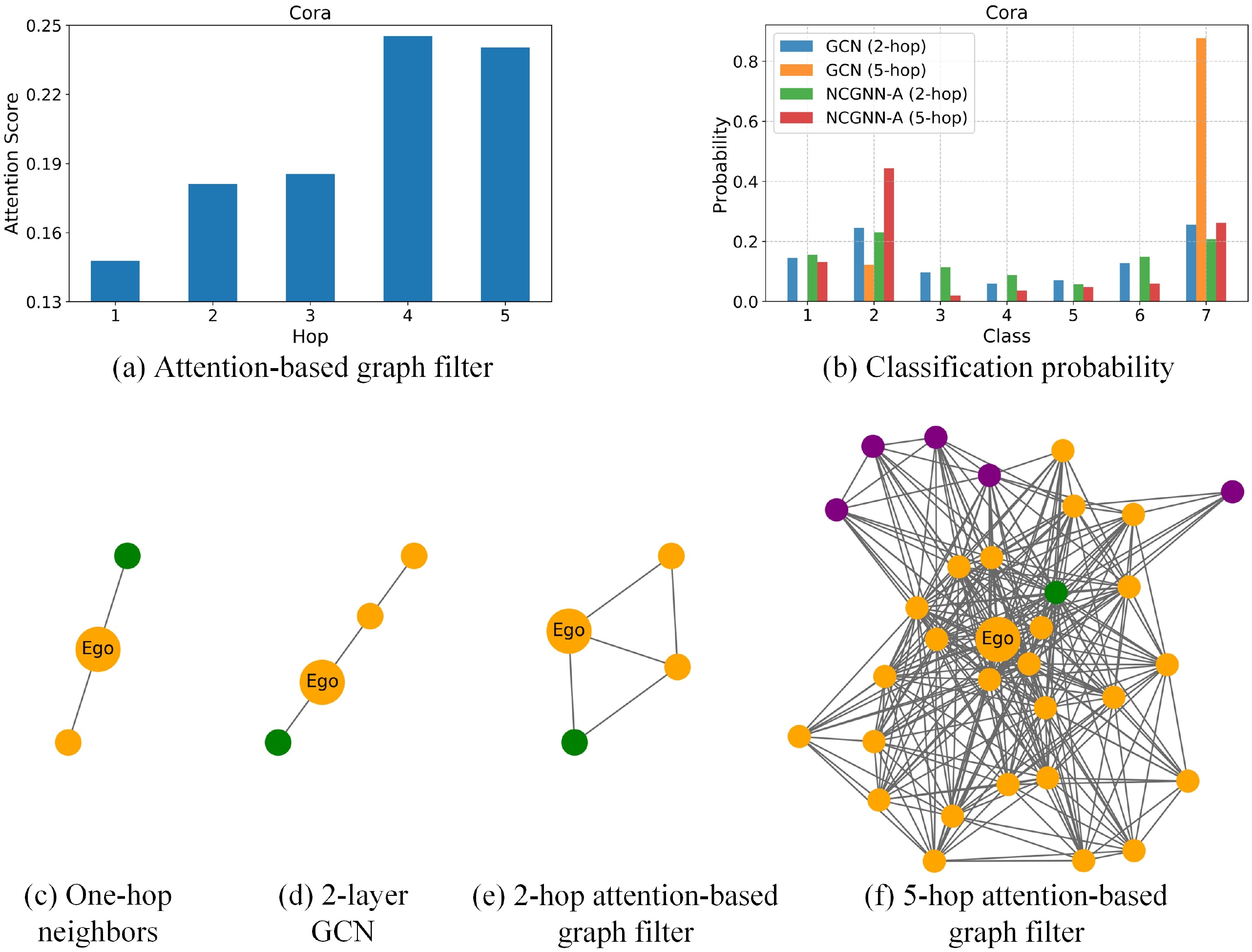}}
\caption{Visualization analysis of the attention-based graph filter on the Cora dataset. (a) The attention scores of different hops of neighborhoods. (b) Classification results on the selected target node, where GCN misclassifies it as the seventh class while NCGNN-A correctly classifies it as the second class. (c-f) Subgraphs of ego networks of the target node extracted by $\mathbf{A}$, GCN (2-hop), NCGNN-A (2-hop), and NCGNN-A (5-hop), respectively. Different colors denote different classes.\label{fig5}}
\end{figure}

We first analyze the attention-based graph filter for identifying the subgraph. As shown in Fig.~\ref{fig5}, larger attention scores are assigned to higher-order neighborhoods, which also verifies that NCGNN-A is indeed exempt from the over-smoothing problem in extracting long-range dependencies. Moreover, we take a close look at the selected target node. Taking one node from Cora that is misclassified by GCN while assigned true label by NCGNN-A as an example, we visualize the receptive field (a subgraph of ego network) of NCGNN-A with $\max(\mathcal{M})=5$ and compare it with the 2-hop neighborhood. For a better view, we only plot the top-32 neighbors in Fig.~\ref{fig5}(f). We find that (i) the 2-hop receptive field of GCN is too small to aggregate adequate node features for correctly classifying the target node, (ii) GCN (5-hop) exaggerates the classification error and suffers from the over-smoothing issue, and (iii) NCGNN-A (5-hop) correctly classifies the target node with the largest probability, verifying that higher-order and more relevant messages from neighboring nodes can be extracted by the 5-hop attention-based graph filter to help NCGNN-A learn effective node representation for classification.

\begin{figure*}[t]
\centerline{\includegraphics[scale=0.266]{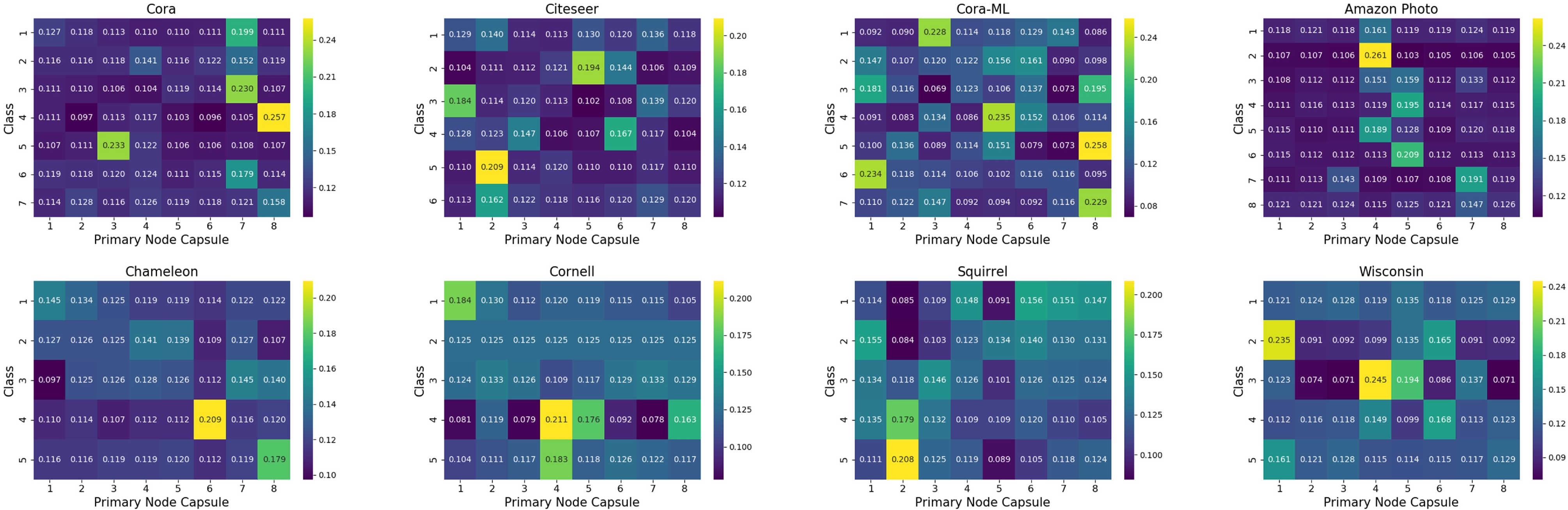}}
\caption{Heatmaps of the matrices of average coupling coefficients obtained by NCGNN-A on eight benchmark datasets.\label{fig6}}
\end{figure*}

Furthermore, we provide the heatmaps of average coupling coefficients (i.e., $\frac{1}{N}\sum_{j=1}^{N}c_{jkl}$) adaptively computed on eight benchmark datasets in Fig.~\ref{fig6} to visualize the average contributions of the input primary node capsules to each class. As shown in~\cite{inter_caps} that interprets capsule-based networks, the coupling coefficient $c_{jkl}$ can be regarded as the contribution of the input node capsule $\mathbf{h}_{j}^{(k)}$ to the capsule $\mathbf{p}_j^{(l)}$ associated with the properties of class $l$ extracted from node $v_j$. Here, we set the number of primary node capsules to 8 for all datasets. The brightness values in each row of the coupling coefficient matrices measure the average influence of a subset of important primary node capsules on the corresponding class. For example, in the Amazon Photo dataset, the fourth primary node capsule is most influential for predicting the node as the second class in average. Thus, these adaptively computed coupling coefficients make the node-level capsule-based message passing transparent in interpreting which part of input node capsules contribute most to the model's prediction.

\subsubsection{Robustness Analysis}
The message passing scheme of our capsule graph layer is able to adaptively restrain noisy features from neighborhoods. In this section, we study the robustness of our method under the missing edge scenarios, where we perturb graphs by randomly removing a certain fraction of edges of the original graph. We use two synthetic graphs with heterophily ($H(\mathcal{G})=0.3$) and homophily ($H(\mathcal{G})=0.7$), respectively, to evaluate the model robustness. 

\begin{figure}[t]
\centerline{\includegraphics[scale=0.3]{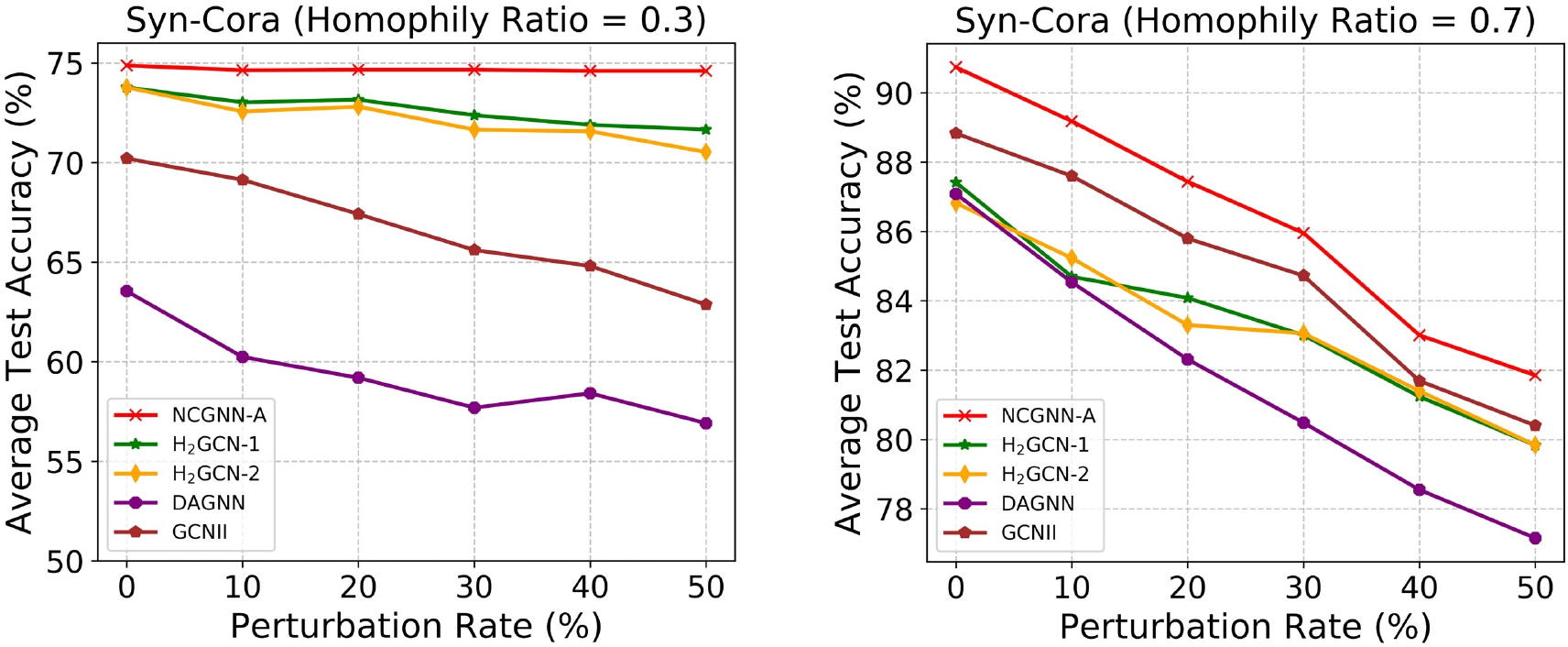}}
\caption{Robustness analysis of NCGNN under both heterophily and homophily.}
\label{fig7}
\end{figure}

We compare with two strong baselines for assortative graphs (i.e., DAGNN and GCNII) and the state-of-the-art method for disassortative graphs (i.e., $\text{H}_{\text{2}}\text{GCN}$). Fig.~\ref{fig7} presents the classification results of these methods in the incomplete graph scenarios, where the results are obtained by averaging over 10 random weight initializations for each perturbation rate. We observe that our NCGNN consistently outperforms all the baselines across all perturbation rates on both assortative and disassortative graphs and demonstrates surprising robustness for learning over the disassortative graph. $\text{H}_{\text{2}}\text{GCN}$ is also robust under heterophily but is outperformed by our method in all cases. DAGNN and GCNII perform poorly under heterophily and do not demonstrate robustness with varying edge perturbations under homophily.

\section{Conclusions} \label{sec5}
In this paper, we propose Node-level Capsule Graph Neural Network (NCGNN) for semisupervised node classification. NCGNN represents nodes as groups of capsules. By leveraging the designed graph filter and capsule graph layer, each node-level capsule adaptively aggregates a subset of advantageous capsules and restrains irrelevant messages from multi-hop neighborhoods. This novel message passing scheme is demonstrated to overcome feature over-mixing to produce better node representations. Experiments on twelve real-world graphs as well as synthetic graphs show that NCGNN performs well under both homophily and heterophily and is exempt from the over-smoothing issue. Moreover, it is inherently interpretable by identifying a subset of node features that are crucial for the model's prediction from the extracted subgraph.

\ifCLASSOPTIONcaptionsoff
\newpage
\fi

\bibliographystyle{IEEEtran}
\bibliography{IEEEabrv,./reference}

\begin{thebibliography}{10}
\providecommand{\url}[1]{#1}
\csname url@samestyle\endcsname
\providecommand{\newblock}{\relax}
\providecommand{\bibinfo}[2]{#2}
\providecommand{\BIBentrySTDinterwordspacing}{\spaceskip=0pt\relax}
\providecommand{\BIBentryALTinterwordstretchfactor}{4}
\providecommand{\BIBentryALTinterwordspacing}{\spaceskip=\fontdimen2\font plus
\BIBentryALTinterwordstretchfactor\fontdimen3\font minus
  \fontdimen4\font\relax}
\providecommand{\BIBforeignlanguage}[2]{{%
\expandafter\ifx\csname l@#1\endcsname\relax
\typeout{** WARNING: IEEEtran.bst: No hyphenation pattern has been}%
\typeout{** loaded for the language `#1'. Using the pattern for}%
\typeout{** the default language instead.}%
\else
\language=\csname l@#1\endcsname
\fi
#2}}
\providecommand{\BIBdecl}{\relax}
\BIBdecl

\bibitem{gcn}
T.~N. Kipf and M.~Welling, ``Semi-supervised classification with graph
  convolutional networks,'' in \emph{5th Int. Conf. Learn. Rep.}, Toulon,
  France, Apr. 2017.

\bibitem{lgcn}
H.~Gao, Z.~Wang, and S.~Ji, ``Large-scale learnable graph convolutional
  networks,'' in \emph{Proc. 24th ACM SIGKDD Int. Conf. Knowl. Disc. Data
  Min.}, London, UK, Aug. 2018, pp. 1414--1426.

\bibitem{gat}
P.~Veli{\v{c}}kovi{\'c}, G.~Cucurull, A.~Casanova, A.~Romero, P.~Lio, and
  Y.~Bengio, ``Graph attention networks,'' in \emph{6th Int. Conf. Learn.
  Rep.}, Vancouver, BC, Canada, Apr. 2018.

\bibitem{ecc}
M.~Simonovsky and N.~Komodakis, ``Dynamic edge-conditioned filters in
  convolutional neural networks on graphs,'' in \emph{Proc. IEEE Comput. Soc.
  Conf. Comput. Vis. Pattern Recognit.}, Honolulu, HI, USA, Jul. 2017, pp.
  3693--3702.

\bibitem{edgeconv}
Y.~Wang, Y.~Sun, Z.~Liu, S.~E. Sarma, M.~M. Bronstein, and J.~M. Solomon,
  ``Dynamic graph {CNN} for learning on point clouds,'' \emph{ACM Trans.
  Graphics}, vol.~38, no.~5, pp. 1--12, Nov. 2019.

\bibitem{molecule}
D.~K. Duvenaud \emph{et~al.}, ``Convolutional networks on graphs for learning
  molecular fingerprints,'' in \emph{Adv. Neural Inf. Process. Syst. 28},
  Montr{\'{e}}al, QC, Canada, Dec. 2015, pp. 2224--2232.

\bibitem{mpnn}
J.~Gilmer, S.~S. Schoenholz, P.~F. Riley, O.~Vinyals, and G.~E. Dahl, ``Neural
  message passing for quantum chemistry,'' in \emph{Proc. 34th Int. Conf. Mach.
  Learn.}, Sydney, NSW, Australia, Aug. 2017, pp. 1263--1272.

\bibitem{recom}
R.~Ying, R.~He, K.~Chen, P.~Eksombatchai, W.~L. Hamilton, and J.~Leskovec,
  ``Graph convolutional neural networks for web-scale recommender systems,'' in
  \emph{Proc. 24th ACM SIGKDD Int. Conf. Knowl. Disc. Data Min.}, London, UK,
  Aug. 2018, pp. 974--983.

\bibitem{finger}
K.~Zhang, Y.~Zhu, J.~Wang, and J.~Zhang, ``Adaptive structural fingerprints for
  graph attention networks,'' in \emph{8th Int. Conf. Learn. Rep.}, Addis
  Ababa, Ethiopia, Apr. 2020.

\bibitem{edge}
L.~Gong and Q.~Cheng, ``Exploiting edge features for graph neural networks,''
  in \emph{Proc. IEEE Comput. Soc. Conf. Comput. Vis. Pattern Recognit.}, Long
  Beach, CA, USA, Jun. 2019, pp. 9203--9211.

\bibitem{relation}
S.~Vashishth, S.~Sanyal, V.~Nitin, and P.~Talukdar, ``Composition-based
  multi-relational graph convolutional networks,'' in \emph{8th Int. Conf.
  Learn. Rep.}, Addis Ababa, Ethiopia, Apr. 2020.

\bibitem{geomgcn}
H.~Pei, B.~Wei, K.~C.-C. Chang, Y.~Lei, and B.~Yang, ``{Geom-GCN}: Geometric
  graph convolutional networks,'' in \emph{8th Int. Conf. Learn. Rep.}, Addis
  Ababa, Ethiopia, Apr. 2020.

\bibitem{curvature}
Z.~Ye, K.~S. Liu, T.~Ma, J.~Gao, and C.~Chen, ``Curvature graph network,'' in
  \emph{8th Int. Conf. Learn. Rep.}, Addis Ababa, Ethiopia, Apr. 2020.

\bibitem{birds}
M.~McPherson, L.~Smith-Lovin, and J.~M. Cook, ``Birds of a feather: Homophily
  in social networks,'' \emph{Annu. Rev. Sociol.}, vol.~27, no.~1, pp.
  415--444, 2001.

\bibitem{explainer}
Z.~Ying, D.~Bourgeois, J.~You, M.~Zitnik, and J.~Leskovec, ``{GNNExplainer}:
  Generating explanations for graph neural networks,'' in \emph{Adv. Neural
  Inf. Process. Syst. 32}, Vancouver, BC, Canada, Dec. 2019, pp. 9244--9255.

\bibitem{xgnn}
H.~Yuan, J.~Tang, X.~Hu, and S.~Ji, ``{XGNN}: Towards model-level explanations
  of graph neural networks,'' in \emph{Proc. 26th ACM SIGKDD Int. Conf. Knowl.
  Disc. Data Min.}, San Diego, CA, USA, Aug. 2020, pp. 430--438.

\bibitem{deeper}
Q.~Li, Z.~Han, and X.-M. Wu, ``Deeper insights into graph convolutional
  networks for semi-supervised learning,'' in \emph{Proc. 32nd AAAI Conf.
  Artif. Intell.}, New Orleans, LA, USA, Feb. 2018, pp. 3538--3545.

\bibitem{protein}
A.~Fout, J.~Byrd, B.~Shariat, and A.~Ben-Hur, ``Protein interface prediction
  using graph convolutional networks,'' in \emph{Adv. Neural Inf. Process.
  Syst. 30}, Long Beach, CA, USA, Dec. 2017, pp. 6530--6539.

\bibitem{caps4}
G.~E. Hinton, A.~Krizhevsky, and S.~D. Wang, ``Transforming auto-encoders,'' in
  \emph{Int. Conf. Artif. Neural Netw.}, Espoo, Finland, Jun. 2011, pp. 44--51.

\bibitem{caps1}
S.~Sabour, N.~Frosst, and G.~E. Hinton, ``Dynamic routing between capsules,''
  in \emph{Adv. Neural Inf. Process. Syst. 30}, Long Beach, CA, USA, Dec. 2017,
  pp. 3856--3866.

\bibitem{caps2}
G.~E. Hinton, S.~Sabour, and N.~Frosst, ``Matrix capsules with {EM} routing,''
  in \emph{6th Int. Conf. Learn. Rep.}, Vancouver, BC, Canada, May 2018.

\bibitem{caps_equi}
J.~E. Lenssen, M.~Fey, and P.~Libuschewski, ``Group equivariant capsule
  networks,'' in \emph{Adv. Neural Inf. Process. Syst. 31}, Montr{\'{e}}al, QC,
  Canada, Dec. 2018.

\bibitem{caps3}
Y.-H.~H. Tsai, N.~Srivastava, H.~Goh, and R.~Salakhutdinov, ``Capsules with
  inverted dot-product attention routing,'' in \emph{8th Int. Conf. Learn.
  Rep.}, Addis Ababa, Ethiopia, Apr. 2020.

\bibitem{capsule2}
S.~Verma and Z.-L. Zhang, ``Graph capsule convolutional neural networks,'' in
  \emph{Joint ICML and IJCAI Workshop Comp. Biol.}, 2018.

\bibitem{capsule0}
M.~D.~G. Mallea, P.~Meltzer, and P.~J. Bentley, ``Capsule neural networks for
  graph classification using explicit tensorial graph representations,''
  \emph{arXiv preprint arXiv:1902.08399}, 2019.

\bibitem{capsule1}
Z.~Xinyi and L.~Chen, ``Capsule graph neural network,'' in \emph{7th Int. Conf.
  Learn. Rep.}, New Orleans, LA, USA, May 2019.

\bibitem{disen}
J.~Ma, P.~Cui, K.~Kuang, X.~Wang, and W.~Zhu, ``Disentangled graph
  convolutional networks,'' in \emph{Proc. 36th Int. Conf. Mach. Learn.}, Long
  Beach, CA, USA, Jun. 2019, pp. 4212--4221.

\bibitem{caps2ne}
D.~Q. Nguyen, T.~D. Nguyen, D.~Q. Nguyen, and D.~Phung, ``A capsule
  network-based model for learning node embeddings,'' in \emph{Proc. 29th ACM
  Int. Conf. Inf. Knowl. Manage.}, Virtual Event, Ireland, Oct. 2020, pp.
  3313--3316.

\bibitem{hgcn}
J.~Yang \emph{et~al.}, ``Hierarchical graph capsule network,'' in \emph{Proc.
  36th AAAI Conf. Artif. Intell.}, Virtual, Feb. 2021, pp. 10\,603--10\,611.

\bibitem{beyond}
J.~Zhu, Y.~Yan, L.~Zhao, M.~Heimann, L.~Akoglu, and D.~Koutra, ``Beyond
  homophily in graph neural networks: Current limitations and effective
  designs,'' in \emph{Adv. Neural Inf. Process. Syst. 33}, Virtual, Dec. 2020,
  pp. 7793--7804.

\bibitem{collective}
P.~Sen, G.~Namata, M.~Bilgic, L.~Getoor, B.~Galligher, and T.~Eliassi-Rad,
  ``Collective classification in network data,'' \emph{AI Mag.}, vol.~29,
  no.~3, pp. 93--106, Sep. 2008.

\bibitem{query}
G.~Namata, B.~London, L.~Getoor, and B.~Huang, ``Query-driven active surveying
  for collective classification,'' in \emph{ICML 2012 Workshop Min. Learn.
  Graphs}, Edinburgh, Scotland, UK, Jul. 2012.

\bibitem{struc2vec}
L.~F.~R. Ribeiro, P.~H.~P. Saverese, and D.~R. Figueiredo, ``struc2vec:
  Learning node representations from structural identity,'' in \emph{Proc. 23rd
  ACM SIGKDD Int. Conf. Knowl. Disc. Data Min.}, Halifax, NS, Canada, Aug.
  2017, pp. 385--394.

\bibitem{measuring}
D.~Chen, Y.~Lin, W.~Li, P.~Li, J.~Zhou, and X.~Sun, ``Measuring and relieving
  the over-smoothing problem for graph neural networks from the topological
  view,'' in \emph{Proc. 34th AAAI Conf. Artif. Intell.}, New York, NY, USA,
  Feb. 2020, pp. 3438--3445.

\bibitem{jknet}
K.~Xu, C.~Li, Y.~Tian, T.~Sonobe, K.-i. Kawarabayashi, and S.~Jegelka,
  ``Representation learning on graphs with jumping knowledge networks,'' in
  \emph{Proc. 35th Int. Conf. Mach. Learn.}, Stockholm, Sweden, Jul. 2018, pp.
  5453--5462.

\bibitem{dna}
M.~Fey, ``Just jump: Dynamic neighborhood aggregation in graph neural
  networks,'' in \emph{ICLR Workshop Rep. Learn. Graphs Manifolds}, New
  Orleans, LA, USA, Apr. 2019.

\bibitem{pagerank}
J.~Klicpera, A.~Bojchevski, and S.~G{\"u}nnemann, ``Predict then propagate:
  Graph neural networks meet personalized {PageRank},'' in \emph{7th Int. Conf.
  Learn. Rep.}, New Orleans, LA, USA, Apr. 2019.

\bibitem{deepgcns}
G.~Li, M.~Muller, A.~Thabet, and B.~Ghanem, ``{DeepGCNs}: Can {GCNs} go as deep
  as {CNNs}?'' in \emph{2019 IEEE/CVF Int. Conf. Comput. Vis. (ICCV)}, Seoul,
  Korea (South), Oct. 2019, pp. 9267--9276.

\bibitem{gcnii}
M.~Chen, Z.~Wei, Z.~Huang, B.~Ding, and Y.~Li, ``Simple and deep graph
  convolutional networks,'' in \emph{Proc. 37th Int. Conf. Mach. Learn.},
  Virtual, Jul. 2020, pp. 1725--1735.

\bibitem{dagnn}
M.~Liu, H.~Gao, and S.~Ji, ``Towards deeper graph neural networks,'' in
  \emph{Proc. 26th ACM SIGKDD Int. Conf. Knowl. Disc. Data Min.}, San Diego,
  CA, USA, Aug. 2020, pp. 338--348.

\bibitem{scattering}
Y.~Min, F.~Wenkel, and G.~Wolf, ``Scattering {GCN}: Overcoming oversmoothness
  in graph convolutional networks,'' in \emph{Adv. Neural Inf. Process. Syst.
  33}, Dec. 2020.

\bibitem{pairnorm}
L.~Zhao and L.~Akoglu, ``{PairNorm}: Tackling oversmoothing in {GNNs},'' in
  \emph{8th Int. Conf. Learn. Rep.}, Addis Ababa, Ethiopia, Apr. 2020.

\bibitem{dgn}
K.~Zhou, X.~Huang, Y.~Li, D.~Zha, R.~Chen, and X.~Hu, ``Towards deeper graph
  neural networks with differentiable group normalization,'' in \emph{Adv.
  Neural Inf. Process. Syst. 33}, Virtual, Dec. 2020, pp. 4917--4928.

\bibitem{dropedge}
Y.~Rong, W.~Huang, T.~Xu, and J.~Huang, ``{DropEdge}: Towards deep graph
  convolutional networks on node classification,'' in \emph{8th Int. Conf.
  Learn. Rep.}, Addis Ababa, Ethiopia, Apr. 2020.

\bibitem{bbgdc}
A.~Hasanzadeh \emph{et~al.}, ``Bayesian graph neural networks with adaptive
  connection sampling,'' in \emph{Proc. 37th Int. Conf. Mach. Learn.}, Virtual,
  Jul. 2020, pp. 4094--4104.

\bibitem{dropout}
N.~Srivastava, G.~Hinton, A.~Krizhevsky, I.~Sutskever, and R.~Salakhutdinov,
  ``Dropout: A simple way to prevent neural networks from overfitting,''
  \emph{J. Mach. Learn. Res.}, vol.~15, no.~1, pp. 1929--1958, Jun. 2014.

\bibitem{explainability1}
F.~Baldassarre and H.~Azizpour, ``Explainability techniques for graph
  convolutional networks,'' in \emph{ICML 2019 Workshop Learn. Reason.
  Graph-Structured Rep.}, Long Beach, CA, USA, Jun. 2019.

\bibitem{sa}
M.~Gevrey, I.~Dimopoulos, and S.~Lek, ``Review and comparison of methods to
  study the contribution of variables in artificial neural network models,''
  \emph{Ecol. Model.}, vol. 160, no.~3, pp. 249--264, 2003.

\bibitem{gbp}
J.~Springenberg, A.~Dosovitskiy, T.~Brox, and M.~Riedmiller, ``Striving for
  simplicity: The all convolutional net,'' in \emph{3rd Int. Conf. Learn. Rep.
  (workshop track)}, San Diego, CA, USA, May 2015.

\bibitem{lrp}
S.~Bach, A.~Binder, G.~Montavon, F.~Klauschen, K.-R. M{\"u}ller, and W.~Samek,
  ``On pixel-wise explanations for non-linear classifier decisions by
  layer-wise relevance propagation,'' \emph{PlOS ONE}, vol.~10, no.~7, 2015.

\bibitem{saliency}
K.~Simonyan, A.~Vedaldi, and A.~Zisserman, ``Deep inside convolutional
  networks: Visualising image classification models and saliency maps,''
  \emph{arXiv preprint arXiv:1312.6034}, 2013.

\bibitem{cam}
B.~Zhou, A.~Khosla, A.~Lapedriza, A.~Oliva, and A.~Torralba, ``Learning deep
  features for discriminative localization,'' in \emph{Proc. IEEE Comput. Soc.
  Conf. Comput. Vis. Pattern Recognit.}, Las Vegas, NV, USA, Jun. 2016, pp.
  2921--2929.

\bibitem{excitationbackprop}
J.~Zhang, S.~A. Bargal, Z.~Lin, J.~Brandt, X.~Shen, and S.~Sclaroff, ``Top-down
  neural attention by excitation backprop,'' \emph{Int. J. Comput. Vis.}, vol.
  126, no.~10, pp. 1084--1102, 2018.

\bibitem{explainability2}
P.~E. Pope, S.~Kolouri, M.~Rostami, C.~E. Martin, and H.~Hoffmann,
  ``Explainability methods for graph convolutional neural networks,'' in
  \emph{Proc. IEEE Comput. Soc. Conf. Comput. Vis. Pattern Recognit.}, Long
  Beach, CA, USA, Jun. 2019, pp. 10\,772--10\,781.

\bibitem{pgm}
M.~Vu and M.~T. Thai, ``{PGM-Explainer}: Probabilistic graphical model
  explanations for graph neural networks,'' in \emph{Adv. Neural Inf. Process.
  Syst. 33}, Virtual, Dec. 2020, pp. 12\,225--12\,235.

\bibitem{pgexplainer}
D.~Luo \emph{et~al.}, ``Parameterized explainer for graph neural network,'' in
  \emph{Adv. Neural Inf. Process. Syst. 33}, Virtual, Dec. 2020, pp.
  19\,620--19\,631.

\bibitem{rgexplainer}
C.~Shan, Y.~Shen, Y.~Zhang, X.~Li, and D.~Li, ``Reinforcement learning enhanced
  explainer for graph neural networks,'' in \emph{Adv. Neural Inf. Process.
  Syst. 34}, Virtual, Dec. 2021.

\bibitem{xai}
T.~Schnake \emph{et~al.}, ``Higher-order explanations of graph neural networks
  via relevant walks,'' \emph{IEEE Trans. Pattern Anal. Mach. Intell.}, 2021.

\bibitem{subgraphx}
H.~Yuan, H.~Yu, J.~Wang, K.~Li, and S.~Ji, ``On explainability of graph neural
  networks via subgraph explorations,'' in \emph{Proc. 38th Int. Conf. Mach.
  Learn.}, Virtual, Jul. 2021, pp. 12\,241--12\,252.

\bibitem{gem}
W.~Lin, H.~Lan, and B.~Li, ``Generative causal explanations for graph neural
  networks,'' in \emph{Proc. 38th Int. Conf. Mach. Learn.}, 2021, pp.
  6666--6679.

\bibitem{cf}
A.~Lucic, M.~ter Hoeve, G.~Tolomei, M.~de~Rijke, and F.~Silvestri,
  ``{CF-GNNExplainer}: Counterfactual explanations for graph neural networks,''
  in \emph{KDD 2021 Deep Learn. Graphs (DLG) Workshop}, Virtual, Aug. 2021.

\bibitem{rcfgnn}
M.~Bajaj \emph{et~al.}, ``Robust counterfactual explanations on graph neural
  networks,'' in \emph{Adv. Neural Inf. Process. Syst. 34}, Virtual, Dec. 2021.

\bibitem{mgegnn}
X.~Wang, Y.~Wu, A.~Zhang, X.~He, and T.-S. Chua, ``Towards multi-grained
  explainability for graph neural networks,'' in \emph{Adv. Neural Inf.
  Process. Syst. 34}, Virtual, Dec. 2021.

\bibitem{iehgcn}
Y.~Yang, Z.~Guan, J.~Li, W.~Zhao, J.~Cui, and Q.~Wang, ``Interpretable and
  efficient heterogeneous graph convolutional network,'' \emph{{IEEE} Trans.
  Knowl. Data Eng.}, 2021.

\bibitem{inter_caps}
L.~Wang \emph{et~al.}, ``An interpretable deep-learning architecture of capsule
  networks for identifying cell-type gene expression programs from single-cell
  {RNA-sequencing} data,'' \emph{Nat. Mach. Intell.}, vol.~2, no.~11, pp.
  693--703, Nov. 2020.

\bibitem{monet}
F.~Monti, D.~Boscaini, J.~Masci, E.~Rodola, J.~Svoboda, and M.~M. Bronstein,
  ``Geometric deep learning on graphs and manifolds using mixture model
  {CNNs},'' in \emph{Proc. IEEE Comput. Soc. Conf. Comput. Vis. Pattern
  Recognit.}, Honolulu, HI, USA, Jul. 2017, pp. 5425--5434.

\bibitem{independence}
Y.~Liu, X.~Wang, S.~Wu, and Z.~Xiao, ``Independence promoted graph disentangled
  networks,'' in \emph{Proc. 34th AAAI Conf. Artif. Intell.}, New York, NY,
  USA, Feb. 2020, pp. 4916--4923.

\bibitem{mixhop}
S.~Abu-El-Haija \emph{et~al.}, ``{MixHop}: Higher-order graph convolutional
  architectures via sparsified neighborhood mixing,'' in \emph{Proc. 36th Int.
  Conf. Mach. Learn.}, Long Beach, CA, USA, Jun. 2019, pp. 21--29.

\bibitem{motif}
J.~B. Lee, R.~A. Rossi, X.~Kong, S.~Kim, E.~Koh, and A.~Rao, ``Graph
  convolutional networks with motif-based attention,'' in \emph{Proc. 28th ACM
  Int. Conf. Inf. Knowl. Manag.}, Beijing, China, Nov. 2019, p. 499–508.

\bibitem{smallworld}
H.~Mehlhorn and F.~Schreiber, \emph{Small-World Property}.\hskip 1em plus 0.5em
  minus 0.4em\relax New York, NY: Springer New York, 2013, pp. 1957--1959.

\bibitem{gdc}
J.~Klicpera, S.~Wei{\ss}enberger, and S.~G{\"u}nnemann, ``Diffusion improves
  graph learning,'' in \emph{Adv. Neural Inf. Process. Syst. 32}, Vancouver,
  BC, Canada, Dec. 2019, pp. 13\,354--13\,366.

\bibitem{scaling}
A.~Bojchevski \emph{et~al.}, ``Scaling graph neural networks with approximate
  {PageRank},'' in \emph{Proc. 26th ACM SIGKDD Int. Conf. Knowl. Disc. Data
  Min.}, San Diego, CA, USA, Aug. 2020, p. 2464–2473.

\bibitem{automating}
A.~K. McCallum, K.~Nigam, J.~Rennie, and K.~Seymore, ``Automating the
  construction of internet portals with machine learning,'' \emph{Inf. Retr.},
  vol.~3, no.~2, pp. 127--163, Jul. 2000.

\bibitem{gauss}
A.~Bojchevski and S.~G{\"u}nnemann, ``Deep {Gaussian} embedding of graphs:
  Unsupervised inductive learning via ranking,'' in \emph{6th Int. Conf. Learn.
  Rep.}, Vancouver, BC, Canada, May 2018.

\bibitem{pitfall}
O.~Shchur, M.~Mumme, A.~Bojchevski, and S.~G{\"u}nnemann, ``Pitfalls of graph
  neural network evaluation,'' in \emph{Relat. Rep. Learn. Workshop (R2L 2018),
  NeurIPS}, Montr{\'{e}}al, QC, Canada, Dec. 2018.

\bibitem{wiki}
B.~Rozemberczki, C.~Allen, and R.~Sarkar, ``Multi-scale attributed node
  embedding,'' \emph{arXiv preprint arXiv:1909.13021}, 2019.

\bibitem{actor}
J.~Tang, J.~Sun, C.~Wang, and Z.~Yang, ``Social influence analysis in
  large-scale networks,'' in \emph{Proc. 15th ACM SIGKDD Int. Conf. Knowl.
  Disc. Data Min.}, Paris, France, Jun. 2009, pp. 807--816.

\bibitem{sgc}
F.~Wu, A.~Souza, T.~Zhang, C.~Fifty, T.~Yu, and K.~Weinberger, ``Simplifying
  graph convolutional networks,'' in \emph{Proc. 36th Int. Conf. Mach. Learn.},
  Long Beach, CA, USA, Jun. 2019, pp. 6861--6871.

\bibitem{adam}
D.~P. Kingma and J.~Ba, ``Adam: A method for stochastic optimization,'' in
  \emph{3rd Int. Conf. Learn. Rep.}, San Diego, CA, USA, May 2015.

\bibitem{tsne}
L.~van~der Maaten and G.~Hinton, ``Visualizing data using {t-SNE},'' \emph{J.
  Mach. Learn. Res.}, vol.~9, no.~86, pp. 2579--2605, Nov. 2008.

\end{thebibliography}

\end{document}